\documentclass[sigconf]{acmart}

\AtBeginDocument{%
  \providecommand\BibTeX{{%
    \normalfont B\kern-0.5em{\scshape i\kern-0.25em b}\kern-0.8em\TeX}}}

\setcopyright{none}
\acmYear{2018}
\acmDOI{}
\acmConference[Conference acronym 'XX]{Make sure to enter the correct
  conference title from your rights confirmation emai}{June 03--05,
  2018}{Woodstock, NY}
\acmISBN{}

\usepackage{algorithm}
\usepackage{algorithmicx}
\usepackage{algpseudocode}
\usepackage{amsfonts}
\usepackage{amsmath}
\usepackage{algpseudocode}
\usepackage{booktabs}
\usepackage{tabularx}
\usepackage{graphicx}
\usepackage{subfigure}
\usepackage{multirow}
\usepackage{tabularx}
\usepackage{newfloat}
\usepackage{listings}
\usepackage{pifont}
\usepackage{float}
\usepackage{caption}
\usepackage{balance}

\begin{document}

\title[Harvesting Efficient On-Demand Order Pooling from Skilled Couriers]{Harvesting Efficient On-Demand Order Pooling from Skilled Couriers: Enhancing Graph Representation Learning for Refining Real-time
Many-to-One Assignments}

\author{Yile Liang}
\affiliation{%
  \institution{Meituan}
  \streetaddress{Wangjing}
  \city{Beijing}
  \country{China}}
\email{yileliang0412@163.com}

\author{Jiuxia Zhao}
\affiliation{%
  \institution{Meituan}
  \streetaddress{Wangjing}
  \city{Beijing}
  \country{China}}
\email{zhaojiuxia@meituan.com}

\author{Donghui Li}
\affiliation{%
  \institution{Meituan}
  \streetaddress{Wangjing}
  \city{Beijing}
  \country{China}}
\email{lidonghui03@meituan.com}

\author{Jie Feng}
\authornote{Corresponding author.}
\affiliation{%
  \institution{Tsinghua University}
  \streetaddress{Wangjing}
  \city{Beijing}
  \country{China}}
\email{fengj12ee@hotmail.com}

\author{Chen Zhang}
\authornote{This work was fulfilled when Chen Zhang interned at Meituan.}
\affiliation{%
  \institution{Tsinghua University}
  \streetaddress{Wangjing}
  \city{Beijing}
  \country{China}}
\email{zhangchen0715@gmail.com}

\author{Xuetao Ding}
\affiliation{%
  \institution{Meituan}
  \streetaddress{Wangjing}
  \city{Beijing}
  \country{China}}
\email{dingxuetao@meituan.com}

\author{Jinghua Hao}
\affiliation{%
  \institution{Meituan}
  \streetaddress{Wangjing}
  \city{Beijing}
  \country{China}}
\email{haojinghua@meituan.com}

\author{Renqing He}
\affiliation{%
  \institution{Meituan}
  \streetaddress{Wangjing}
  \city{Beijing}
  \country{China}}
\email{herenqing@meituan.com}

\renewcommand{\shortauthors}{Yile Liang et al.}

\begin{abstract}
 The recent past has witnessed a notable surge in on-demand food delivery (OFD) services, offering delivery fulfillment within dozens of minutes after an order is placed.  In OFD, pooling multiple orders for simultaneous delivery in real-time order assignment is a pivotal efficiency source, which may in turn extend delivery time. Constructing high-quality order pooling to harmonize platform efficiency with the experiences of consumers and couriers, is crucial to OFD platforms. 
 However, the complexity and real-time nature of order assignment, making extensive calculations impractical, significantly limit the potential for order consolidation.
 Moreover, offline environment is frequently riddled with unknown factors, posing challenges for the platform's perceptibility and pooling decisions. 
 
 Nevertheless, delivery behaviors of skilled couriers (SCs) who know the environment well, can improve system awareness and effectively inform decisions. 
 Hence a SC delivery network (SCDN) is constructed, based on an enhanced attributed heterogeneous network embedding approach tailored for OFD. It aims to extract features from rich temporal and spatial information, and uncover the latent potential for order combinations embedded within SC trajectories.
 Accordingly, the vast search space of order assignment can be effectively pruned through scalable similarity calculations of low-dimensional vectors, making comprehensive and high-quality pooling outcomes more easily identified in real time. 
 In addition, the acquired embedding outcomes highlight promising subspaces embedded within this space, i.e., scale-effect hotspot areas, which can offer significant potential for elevating courier efficiency. 
 
 SCDN has now been deployed in Meituan dispatch system. Online tests reveal that with SCDN, the pooling quality and extent have been greatly improved. And our system can boost couriers' efficiency by 45-55\% during noon peak hours, while upholding the timely delivery commitment. 
\end{abstract}


\maketitle

\section{Introduction}
\subsection{Backgrounds}
In recent years, there has been a remarkable upsurge in the widespread adoption of on-demand food delivery (OFD) services world-wide. 
With a mere few clicks, consumers can enjoy delicious meals without stepping out, all delivered right to their doorstep within just a few dozen minutes.
This trend is attributable to the overarching shifts in technological innovation, including the popularity of apps and online platforms, and the growing dependence on third-party services for OFD. 
Global revenues for OFD sector were about \$90 billion in 2018, rose to \$294 billion in 2021, and are expected to exceed \$466 billion by 2026 \cite{meemken2022research}.
Meituan Waimai, China's pioneering OFD platform has witnessed remarkable growth over the last decade. In 2023, the platform handles over 70 million orders daily, encompassing an extensive reach across almost 3,000 cities, counties and regions throughout China.
6.24 million couriers earned income via Meituan, with over 1 million actively engaged daily. 

In OFD, orders are placed continuously by consumers from various locations. In response, the platform promptly gathers these newly initiated orders, channels them to merchants, and assigns dedicated couriers for pick-up and delivery within the promised delivery time. The platforms act as intermediaries, linking a multitude of consumers, merchants and couriers within the ecosystem, and strike a balance between gains and losses among these stakeholders to achieve sustained growth and prosperity \cite{liang2023enhancing}. Among these, consumers desire prompt services, merchants seek to maintain food freshness, couriers aim to fulfill enough orders to earn a decent income in a safe environment, while OFD platforms focus on boosting efficiency to reduce costs and increase profits.

\begin{figure}[!ht]
	\centering
	\includegraphics[width=1.0\columnwidth]{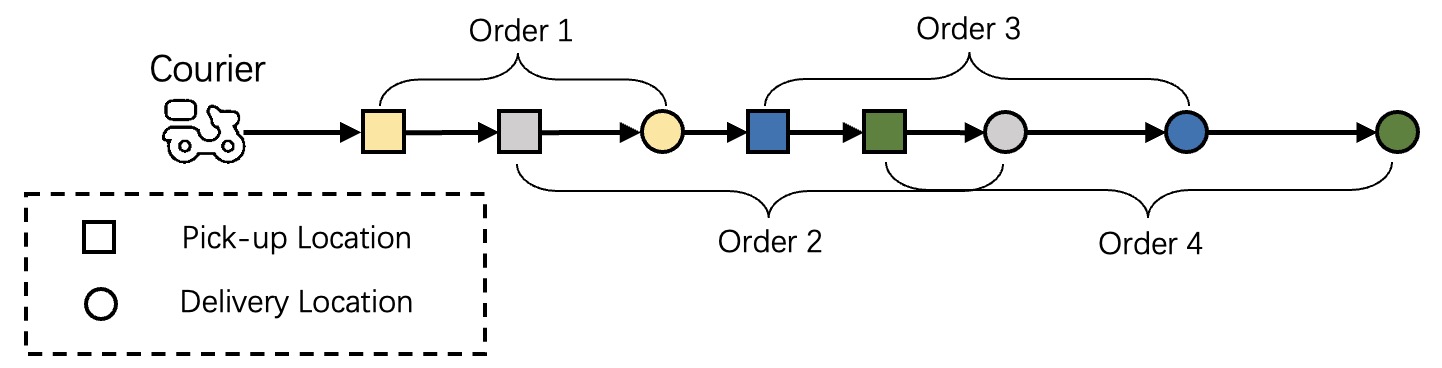}
	\caption{A courier's concurrent execution and route sequence of four orders.\label{fig:courier_concurrent}}
\end{figure}

\begin{figure}[!ht]
  \centering
  \subfigure[High-quality order pooling.]{
    \label{fig:good_route}
    \includegraphics[width=0.85\columnwidth]{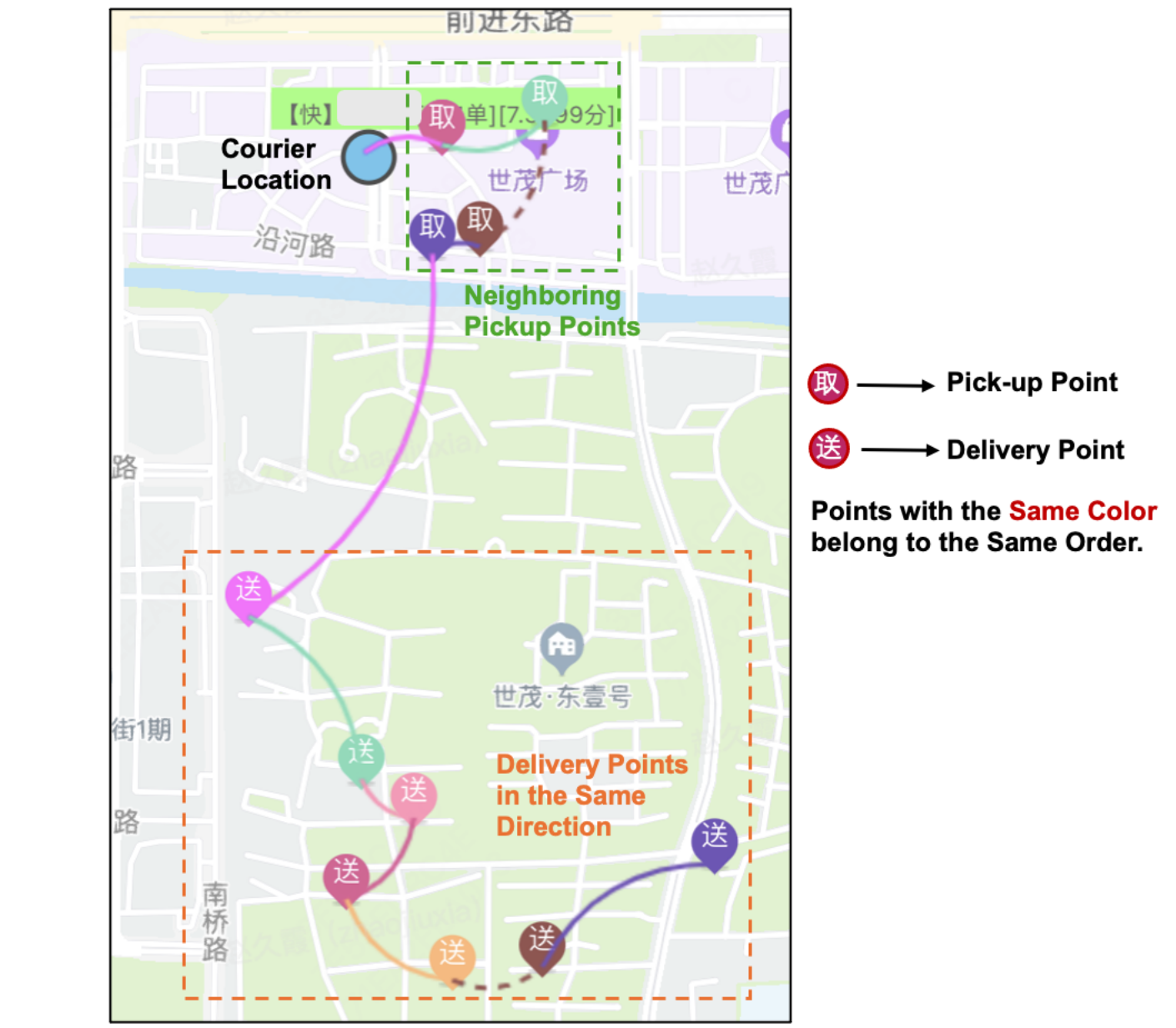}
  }
  \subfigure[Unreasonable order pooling.]{
    \label{fig:bad_route}
    \includegraphics[width=0.7\columnwidth]{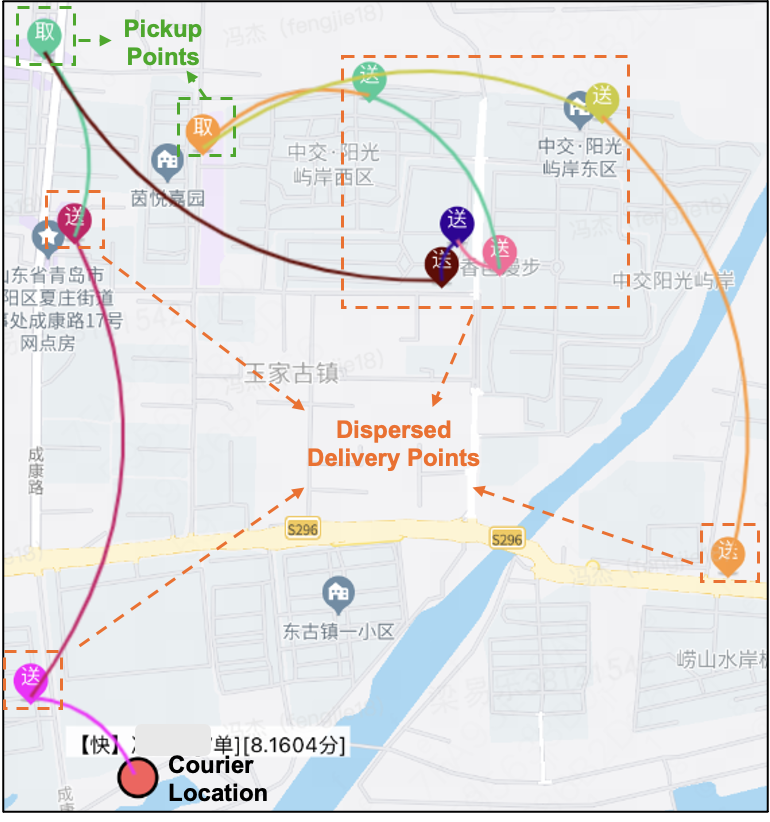}
  }
  \caption{Order pooling examples.}
\end{figure}

In this context, couriers often engage in concurrent execution of multiple delivery tasks, including order pick-up and delivery. \textbf{A pivotal efficiency source in OFD is to pool multiple orders for simultaneous delivery of a single courier in order assignment,} leveraging shared pick-up and delivery behaviours and travelling distances, 
enabling couriers serve more orders within committed delivery time limits. Facilitating comprehensive order pooling can effectively reduce delivery costs and enhance OFD sustainability\cite{shetty2022value, simoni2023crowdsourced}. Figure \ref{fig:good_route} presents a high-quality order pooling example, where the courier's pickup points are highly concentrated, and the delivery destinations are aligned along a coherent route, enabling the courier to fulfill the deliveries with remarkable efficiency.
However, unreasonable order pooling may result in detours and prolonged delivery times, severely undermining the stakeholders' experiences. Figure \ref{fig:bad_route}  illustrates a scenario in which unreasonable order pooling negatively impacts a courier's route, leading to an inefficient delivery trajectory.

In Meituan Waimai, the dispatch system conducts city-level batch order assignments every 30 seconds\cite{liang2023enhancing}. 
In each dispatch cycle, the system identifies available couriers for new orders, and assesses the matching degree (MD) between them, including convenience of route, over-time risk, and courier acceptance willingness. This evaluation process demands massive computations for pick-up and delivery route planning (PDRP) to simulate courier's behaviors after accepting orders \cite{feng2023ilroute}. 
Subsequently, through the resolution of a multi-objective many(order)-to-one(courier) assignment (MOA) problem, the system matches orders with the most suitable couriers to optimize the overall MD scores. 

\textbf{{Constructing comprehensive and high-quality order pooling in order assignments stands as a key issue for OFD platforms}} to harmonize platform efficiency with stakeholder experience.
Practically, there are two primary methods to facilitate comprehensive and high-quality order pooling in order assignments during each dispatch cycle. The first approach entails identifying suitable order combinations among all the pending orders, such as those with shared pick-up/delivery tasks or minimal detours, aiming to increase the ratio of MOA outcomes. The second approach focuses on matching orders with couriers whose existing assignments can share pick-up/delivery tasks or travel routes with the new orders, thereby optimizing the delivery process.

\subsection{Challenges}
However, OFD's distinct features present considerable challenges.

\textbf{(1) Computational complexity in real time.}
On one hand, the MD scores based on PDRP outcomes, are \textbf{\textit{non-additive}}. Specifically, the MD score of assigning multiple orders concurrently to a courier, is not equivalent to the sum of the scores of assigning each order individually to the same courier. 
Hence, to model the MOA problem and to obtain sufficient order combination results usually demands massive MD score calculations, which suffers from combinatorial explosion, as depicted in Figure \ref{fig:challenge_OFD_1}. The MOA problem details can be found in Appendix \ref{APP-MOA}.
For some big cities in China during noon peak, there amounts to over 3 thousand orders \footnote{It is the order volume in several geographically adjacent areas within a city, not the total order volume for the entire city.} to be assigned in each dispatch cycle, while each order can retrieve hundreds of couriers available for delivery on average. Assuming at most 5 orders assigned to a courier, and the average courier candidates for a order (combination) is 100, the calculation volume is $(C_{3000}^1+C_{3000}^2+C_{3000}^3+C_{3000}^4+C_{3000}^5) \times 100$.
On the other hand, the MOA problem itself is categorized as an \textbf{\textit{NP-hard}} integer programming problem, known for its extremely vast search space. Crafting online algorithms that perform effectively for the MOA is an exceptionally challenging task\cite{maniezzo2021matheuristics,zhou2020two,chen2022imitation}.
Moreover, the fast movement of couriers requires assignment decisions be made within a mere 10 seconds. This imperative time frame ensures the consistency of courier status between the information acquisition phase and the actual assignment moment. 

Consequently, the platform tends to favor one(order)-to-one(courier) assignments during each dispatch cycle, a strategy that reduces computational volume and complexity, albeit at the expense of comprehensive order pooling.

\begin{figure}[!ht]
	\centering
	\includegraphics[width=1.0\columnwidth]{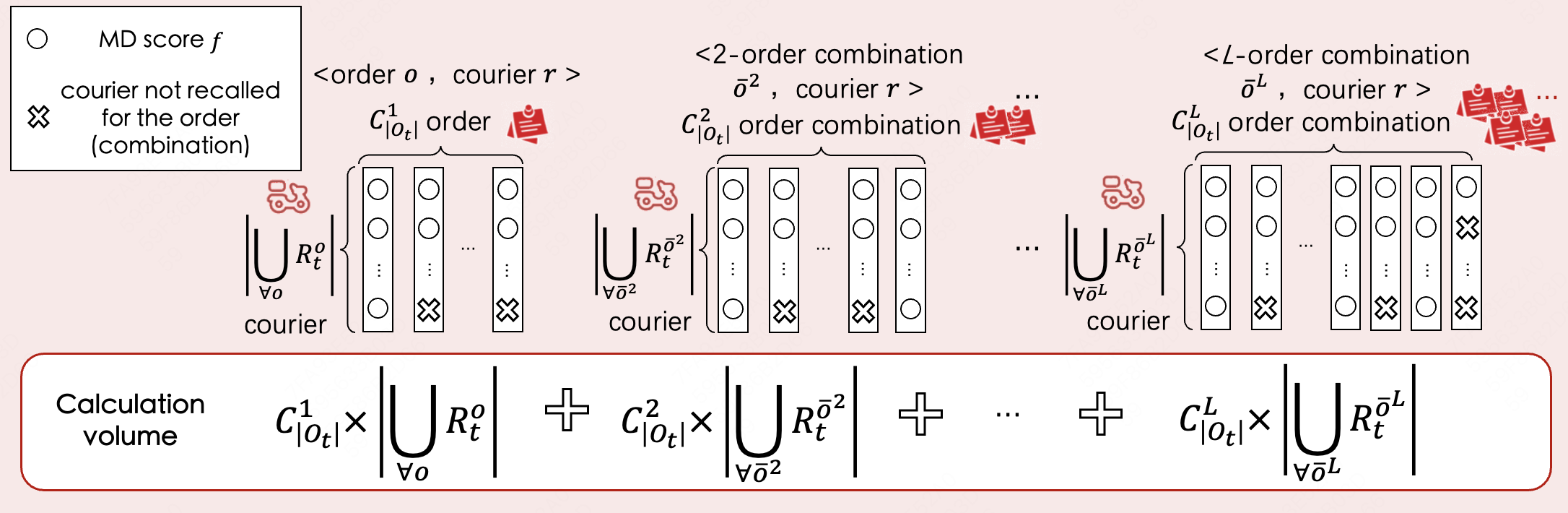}
	\caption{Calculation volume and search space for modeling and solving MOA problems in each dispatch cycle.\label{fig:challenge_OFD_1}}
\end{figure}

\textbf{(2) Limited system awareness on the ``last mile" offline environment.} 
In OFD, the "last mile" offline environment is highly intricate and dynamic \cite{zheng2022solving}, encompassing unforeseen road closures, unknown natural obstacles, and pandemic-related lockdowns. OFD platforms are unable to fully access these extensive, finely-detailed spatiotemporal data during large-scale decision-making, due to insufficient map precision and digital capabilities, along with computational and storage constraints. 
Consequently, order pooling decisions based on coarse data and limited awareness, may not be reasonable, potentially harming courier experiences, causing delivery delays, and reducing delivery efficiency.

\subsection{Related Work}
Prior research on order pooling algorithms primarily focused on batching issues in traditional warehouse management \cite{pardo2023order,briant2020efficient,yang2021hybrid}. 
However, the more relaxed time constraints of warehouse batching algorithms, typically in minutes, or even hours, are not well-suitable for the urgency required in OFD.

In recent years, research pertaining to OFD has gradually gained traction.
The prevalent method for order pooling batches orders based on geographical proximity and closeness of their promised delivery time \cite{reyes2018meal}. However, while these criteria-based batching rules are straightforward, they limit the scope for consolidation.
An exact algorithm for order batching and assignment is proposed in \cite{yildiz2019provably}, under the unrealistic assumption of perfect information about the arrival of orders. 
The study in \cite{ji2019alleviating} produces monthly OFD task groupings offline to facilitate order consolidation, However, their effectiveness is heavily reliant on order structure stability. 
Work in \cite{joshi2021batching, joshi2022foodmatch} achieve order consolidation using iterative clustering on an order graph, but the batching algorithm's complexity and computational load hinder real-time processing.
Similar work in \cite{simoni2023crowdsourced} leverages additional decomposition mechanisms to reduce computational cost, yet it falls short of enabling real-time application despite notable performance gains.
To satisfy the need for solutions within seconds, XGBoost models are built through supervised learning on historical order assignment results in \cite{wang2021solving, yu2021delay}, to promote combined order assignments. However, the consolidation results struggle to break through the constraints of historical decisions, resulting in limited effectiveness.
\subsection{Motivations}
In light of the limitations present in existing work, it's worth noting that OFD platforms are equipped with a vast fleet of couriers, and extensive data on courier behaviors, especially from the skilled ones, which offer insights for high-efficiency and quality delivery services and enhance system intelligence.
Skilled couriers (SCs) often possess a comprehensive grasp of the offline environment, including order distribution and road logistics, and continually improve their delivery skills to adapting to complex conditions. Moreover, our couriers can reject or transfer system-assigned orders, leveraging their expertise to optimize routes, minimizing detours and overtime.
Additionally, the platform gathers courier preferences for pick-up and delivery locations via their apps, promoting efficient operations with fewer bottlenecks.
\textbf{Thus SCs' behaviours of order selection, route sequence and feedback can provide the system superior courier-oriented pooling outcomes and help improve decision quality.
}

In the past decade, the work on word representation learning has achieved cutting-edge results \cite{mikolov2013distributed, perozzi2014deepwalk, tang2015line, grover2016node2vec}.
Neural language models replace traditional high-dimensional and sparse word vectors with low-dimensional and dense embeddings, which assume that frequently co-occurring words share stronger statistical dependencies.
Recently, graph representation learning (GRL) methods \cite{cui2018survey, khoshraftar2022survey} have increasingly been applied in various fields, including e-commerce \cite{wang2018billion, hu2016heterosales,grbovic2018real}, job search \cite{kenthapadi2017personalized, ramanath2018towards}, ride-sharing \cite{tang2020efficient, tang2021will}, to discover diverse types of recommendations on the Web. These approaches have had a major impact in both academia and industry. %

Drawing on prior achievements and the principle that \textbf{orders frequently combined together in SCs' routes tend to yield top-tier pooling results}, this paper aims to \textbf{using GRL methods to uncover the latent potential for order pooling embedded within the SCs' behaviour data}. 
Therefore, through scalable low-dimension vector calculations, instead of massive and time-consuming PDRP computations, 
we effectively prune the MOA problem's search space, shown in Figure \ref{fig:challenge_OFD_1}, 
meanwhile extract small-scale and isolated subspaces promising for high-quality order consolidation results, facilitating real-time, effective order pooling.
\subsection{Contributions}
Accordingly, a systemic solution framework, named as SC delivery network (SCDN), is proposed. The novel contributions are:

\textbf{(1) Graph Modelling: } 
We construct a delivery network from SC route sequences, with \textit{flow unit} (FU) as nodes linked by SC behavior sequences. An FU is a directed vector from pick-up areas of interest (AOI \footnote{AOIs are defined as non-overlapping irregular polygons that comprehensively divide and cover the space})\cite{zhu2023c} to delivery AOI. Orders of an FU share the same pick-up and delivery AOIs. 
The network is formulated as an \textbf{\textit{attributed multiplex heterogeneous network}} (AMHEN), 
with FU nodes featuring multiple attributes for temporal and spatial information, and links representing two different types of courier behaviors, namely pick-up and delivery.

\textbf{(2) Learning Algorithm: } Based on GATNE \cite{cen2019representation}, an effective GRL method for AMHEN, an enhanced attributed heterogeneous network embedding (EATNE) approach tailored for OFD is derived to obtain FU embeddings. 
First, given the fact that couriers move within a confined region\footnote{A circular area with a diameter of 3-5 km, and the courier's designated residence as the center.} in a city, a \textbf{{region-congregated negative sampling mechanism}} is proposed as an enhancement over traditional randomized negative sampling to improve algorithm performance. 
Second, we employ a \textbf{customized margin ranking loss} instead of cross-entropy used by GATNE, aiming to refine embedding quality. Last, to address dispersed order distribution and limited FU coverage in SC behaviors, we build a \textbf{cold start mitigation mechanism}, using geographic information to generate embeddings of FUs previously unseen, thus broadening coverage.

\textbf{(3) MOA Search Space Refinement and OFD Application: } 
Utilizing FU embedding, we reconstruct the order combination and courier recall mechanisms within Meituan's dispatch system, facilitating superior real-time order pooling. 
Our use of SCDN refines order structure profiles and pinpoints scale-effect hotspots within MOA's vast search space, uncovering independent and small-scale subspaces for thorough and high-quality order pooling. Accordingly, an innovative delivery mode is developed to enhance courier efficiency without compromising service reliability.

To our knowledge, this is the first application of GRL methods in achieving real-time order pooling in OFD, now deployed in Meituan Waimai's dispatch system. 
Online tests shows significant improvement in order pooling. The total MD score of the MOA problem is improved by 5.3\%, indicating more efficient order assignments with reduced detours and overtime risks. 
The newly-built mode cut the average incremental pick-up time for couriers \footnote{defined as the interval between picking up the current order and the preceding one in the courier's route.} during noon peak by 51\% and delivery time by 21\%.
These enhancements have led to a 45-55\% boost in efficiency, maintaining consistent work hours and on-time delivery standards.

\section{Graph Representation Learning Approach} \label{method}
In this section, we will detail the step-by-step process by which the FU embeddings are acquired.
\subsection{AMHEN Construction}
The AMHEN is constructed based on SC route sequences as described below. The definition of SC and selection criteria of SC route sequences are introduced in Appendix \ref{APP-sc}.

\begin{figure}[!ht]
\captionsetup{justification=justified,singlelinecheck=false}
\centering
\includegraphics[width=1.0\columnwidth]{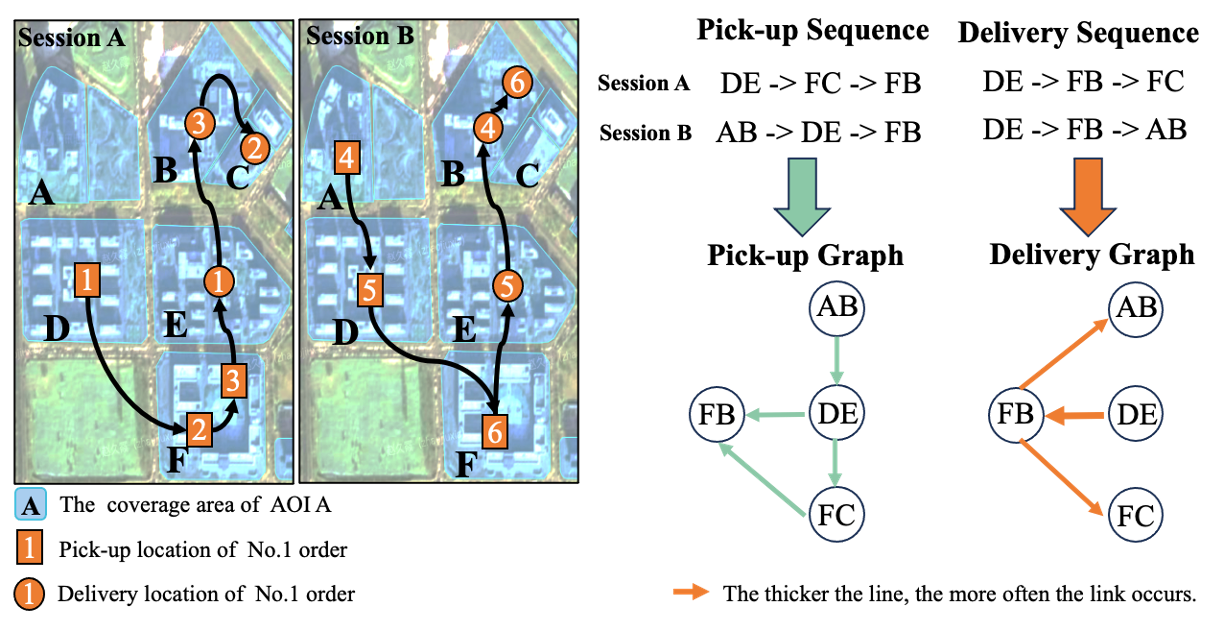}
\caption{Illustration of AMHEN Construction, including 2 sessions. Session A contains 3 orders for FUs DE, FB and FC. The pick-up FU sequence is DE->FC->FB. And the delivery FU sequence is DE->FB->FC. Session B follows the same process.}
\label{fig:AMHEN}
\end{figure}

We first divide a SC's route sequence into distinct \textbf{\textit{sessions}}, using the rest or no action interval as a separator, presently set to 30 minutes. 
Then we transform the route sessions into \textbf{\textit{FU sequences}} via replacing the orders in the sessions with their FUs. Since couriers participate in both pick-up and delivery actions during order fulfilment, there are two kinds of FU sequences: one based on pick-up behavior and the other on delivery, 
as shown in Figure \ref{fig:AMHEN}. 
Diverse couriers' FU sequences may incorporate some common FUs.

To capture shared experiences of SCs, by treating FU as nodes and their connections in the FU sequence as links, we can integrate all the FU sequences into a unified yet \textbf{\textit{heterogeneous graph}}. 
Moreover, it is crucial to utilize the rich temporal and spatial information to enhance learning accuracy, e.g. average historical order amount and delivery distance of each FU, which makes the above graph an \textbf{\textit{AMHEN}}. More about node attributes is in Appendix \ref{APP-node}.

Denote AMHEN by $G = (V, E, A)$, where $V$ is the FU node set, $A$ is the attribute set for all nodes. FU node $v_i \in V$ owns fruitful attributes $\mathbf{x}_i \in A$ to describe its crucial characters.
$E=(E^p, E^d)$ is the set of edges, which contains two types: pick-up and delivery. Specifically, there may be two types of edges between the FU nodes $v_i$ and $v_j$, where $e_{ij}^p \in E^p$ indicates a pick-up edge and $e_{ij}^d \in E^d$ a delivery one.
If two orders, belonging to FU nodes $v_i$ and $v_j$, are successively picked up by the same SC, there exists a pick-up edge $e_{ij}^p$ connecting $v_i$ and $v_j$. 
Similarly, a delivery edge $e_{ij}^d$ indicates there exist orders of FU nodes $v_i$ and $v_j$ that are consecutively delivered by the same SC. Hence, an AMHEN is constructed by merging massive records from tens of thousands of SCs.

\subsection{Graph Representation Learning Model}
\begin{figure}[!ht]
\centering
\includegraphics[width=1.0\columnwidth]{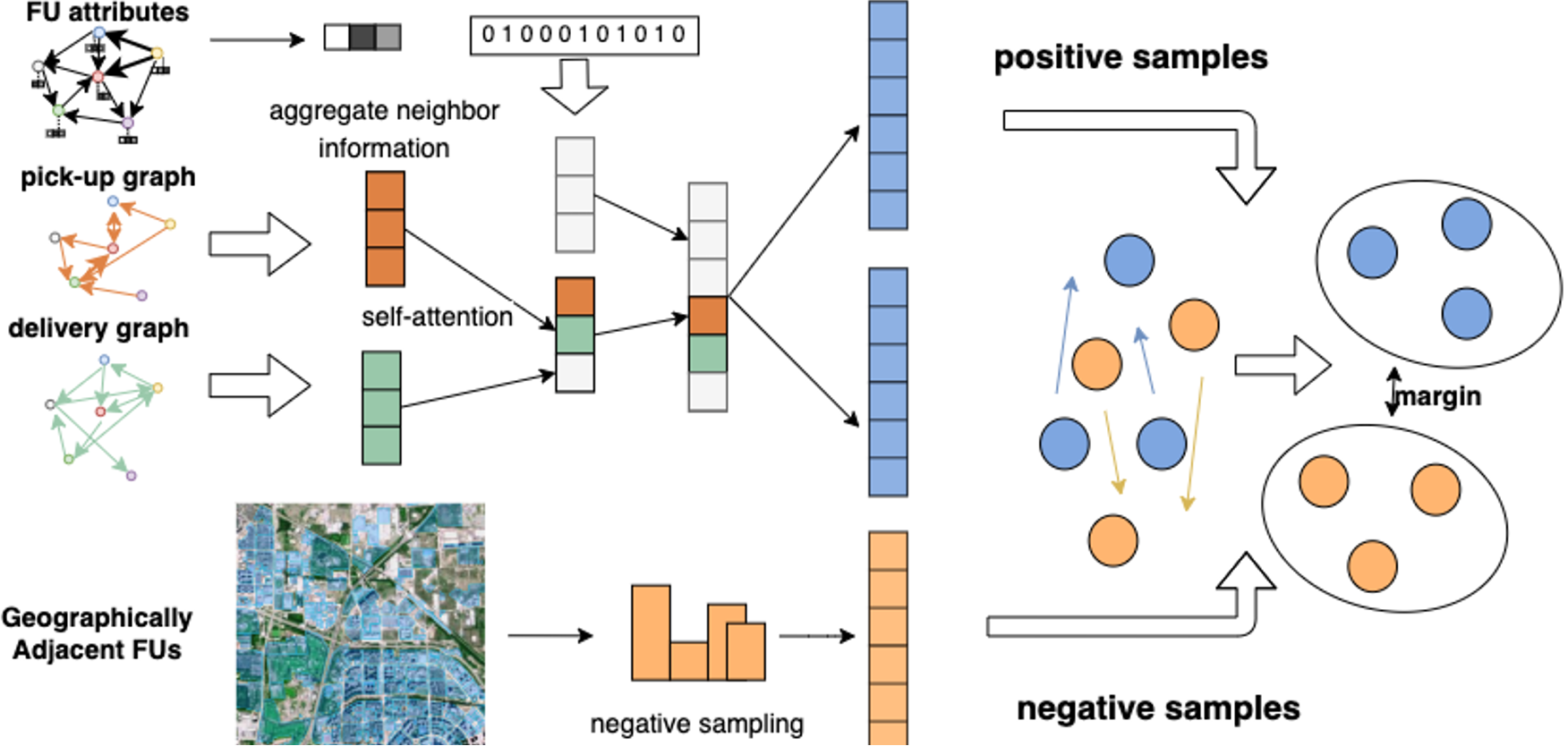}
\caption{Illustration of the GRL Model.}
\label{fig:GRL model}
\end{figure}

Treating the AMHEN as input, we apply the model in GATNE \cite{cen2019representation} to produce node vector representation, i.e. FU embedding,
which can be regarded as the aggregation of various node attributes and topology information in the graph, as depicted in Figure \ref{fig:GRL model}.

We divide the whole embedding of node $v_i$ on each edge type $\tau$ into two parts, base embedding and edge embedding. 
The base embedding $\mathbf{b}_i$ is defined as a parameterized function of its attributes $\mathbf{x}_i$ as $\mathbf{b}_i=\mathbf{h}\left(\mathbf{x}_i\right)$, where $\mathbf{h}$ is a transformation function,
while the k-th level edge embedding $\mathbf{u}_{i, \tau}^{(k)} \in \mathbb{R}^s,(1 \leq k \leq K)$ of node $v_i$ on edge type $\tau$ is aggregated from the edge embeddings of neighbors:
\begin{equation}
\mathbf{u}_{i, \tau}^{(k)}=\text { aggregator }\left(\left\{\mathbf{u}_{j, \tau}^{(k-1)}, \forall v_j \in \mathcal{N}_{i, \tau}\right\}\right),
\end{equation}
\noindent where $\tau \in \{p,d\}$ indicates the edge type,
$s$ is the dimension of edge embeddings,
and $\mathcal{N}_{i, \tau}$ is the neighbors of node $v_i$ on edge type $\tau$. 
The initial edge embedding $\mathbf{u}_{i, \tau}^{(0)}$ is parameterized as the function of attributes $\mathbf{x}_i$ : $\mathbf{u}_{i, \tau}^{(0)}=\mathbf{g}_{\tau}\left(\mathbf{x}_i\right)$, where $\mathbf{g}_{\tau}$ is a transformation function.
The \textit{aggregator} function is mean operation in practice.

We denote the $K$-th level edge embedding $\mathbf{u}_{i, \tau}^{(K)}$ by $\mathbf{u}_{i, \tau}$. Then the pick-up edge embedding $\mathbf{u}_{i, p}$ and the delivery edge embedding $\mathbf{u}_{i, d}$ of node $v_i$ are combined as $\mathbf{U}_{i} = \left(\mathbf{u}_{i, p}, \mathbf{u}_{i, d}\right)$.
Given that the pick-up edge and delivery edge have different impacts, self attention mechanism is used to calculate the weights $\mathbf{{a}}_{i, \tau} \in \left\{ \mathbf{{a}}_{i, p}, \mathbf{{a}}_{i, d}\right\}$.
\begin{equation}
\mathbf{a}_{i,\tau}=\operatorname{softmax}\left(\mathbf{w}_\tau^{\top} \tanh \left(\mathbf{W}_\tau \mathbf{U}_i\right)\right)^{\top},
\end{equation}
where $\mathbf{w}_\tau \in \mathbb{R}^{d_a}, \mathbf{W}_\tau \in \mathbb{R}^{d_a \times s}$ are trainable parameters for edge type $\tau$.
Thus, the overall embedding of node $v_i$ for pick-up edge $\mathbf{v}_{i, p}$ and delivery edge $\mathbf{v}_{i, d}$ can be computed as:
\begin{align}
\mathbf{v}_{i,p}&=\mathbf{h}\left(\mathbf{x}_i\right)+\alpha_p \mathbf{a}_{i, p} \mathbf{M}_p^{\top} \mathbf{u}_{i, p} +\beta_p \mathbf{g}_p \mathbf{x}_i,\label{eq:pick-up}\\
\mathbf{v}_{i,d}&=\mathbf{h}\left(\mathbf{x}_i\right)+\alpha_d \mathbf{a}_{i, d} \mathbf{M}_d^{\top} \mathbf{u}_{i, d} +\beta_d \mathbf{g}_d \mathbf{x}_i,\label{eq:delivery}
\end{align}
where $\alpha_p$ and $\alpha_d$ indicate importance of pick-up and delivery edge embeddings, respectively, characterizing how pick-up and delivery behaviors affect courier efficiency. $\mathbf{M}_p, \mathbf{M}_d \in \mathbb{R}^{s \times d}$ are trainable parameters. $\beta_p$ and $\beta_d$ control the importance of node attributes.

The FU embedding $\mathbf{v}_{i}$ is the average of $\mathbf{v}_{i,p}$ and $\mathbf{v}_{i,d}$. The detailed implementation of EATNE can be found in Appendix \ref{APP-eatne}.
\subsection{Model Optimization}
The positive data for training is generated by a meta-path-based random walk method and skip-gram model \cite{mikolov2013distributed}.
Given a set of pick-up FU sequences $S$, supposing that random walk with length $l$ on $S$ follows a path $S_p=\left(v_{s_1}, \ldots, v_{s_l}\right)$, the pick-up context of $v_{s_t}$ is denoted as $C_{v_{s_t}}^{S_p}=\left\{v_{s_k}\left|v_{s_k} \in S_p,\right| k-t \mid \leq c, t \neq k\right\}$, where $c$ is the size of the sampling window.
Thus, given a node $v_{i}$ and its all pick-up contexts, we can generate a positive pick-up data set $\mathcal{D}_p^P$ of positive pairs ($v_{i}, v_{j}$), which indicates SCs frequently pool the orders of these FU together.
Similarly, we can generate a positive data set $\mathcal{D}_d^P$ from the delivery FU sessions.

\textbf{Negative Sampling.} 
Since couriers usually move within a confined region, negative samples from different regions are so easy for the model to distinguish in the whole training stage which makes the learning inefficient.
Therefore, the negative data sets $\mathcal{D}^N_p$, $\mathcal{D}^N_d$ are constructed by random sampling from pick-up and delivery FU pairs in the same delivery region but excluding positive pairs, respectively. 
In other words, we select k-hop (k>2) neighbors of the FU node that share the same confined region as the challenging negative samples to enable the effective training of the proposed model.
Traditional GATNE uses randomized negative sampling, yet ignores the regional effects in OFD. We find that the performance of GATNE decreases as the negative sampling scope expands and the effect becomes almost random as it reaches the city size.

\textbf{Margin Ranking Loss.} 
The learning task is to make the representation of positive FU pairs lying nearby in the embedding space, and the negative pairs different. However, achieving this with cross-entropy can be challenging.
Therefore, a customized optimization objective based on margin ranking loss is proposed to maximize the distance between positive and negative samples in Equation \ref{eq:obj}, where $\gamma_p^P$, $\gamma_d^P$, $\gamma_p^N$ and $\gamma_d^N$ are hyperparmeters representing the weights of various data sets, $m_p$ and $m_d$ are the minimum distance between negative pairs for pick-up and delivery, and $\cos$ represents the cosine similarity between FU embeddings.

\begin{align}
L &= \frac{\gamma_p^P}{|D_p^P|} \sum_{({v}_{i},{v}_{j}) \in D_p^P}(1-\cos(\mathbf{v}_{i,p},\mathbf{v}_{j,p})) \nonumber \\
&\quad + \frac{\gamma_d^P}{|D_d^P|} \sum_{({v}_{i},{v}_{j}) \in D_d^P}(1-\cos(\mathbf{v}_{i,d},\mathbf{v}_{j,d})) \nonumber \\
&\quad + \frac{\gamma_p^N}{|D^N_p|} \sum_{(v_{i}, v_{j}) \in D^N_p} \max \left(0, \cos(\mathbf{v}_{i,p},\mathbf{v}_{j,p}) - m_p\right) \nonumber \\
&\quad + \frac{\gamma_d^N}{|D^N_d|} \sum_{(v_{i}, v_{j}) \in D^N_d} \max \left(0, \cos(\mathbf{v}_{i,d},\mathbf{v}_{j,d}) - m_d\right), 
\label{eq:obj}
\end{align}

\subsection{Embedding Coverage Improvement} 
\begin{figure}[!ht]
\centering
\includegraphics[width=1\columnwidth]{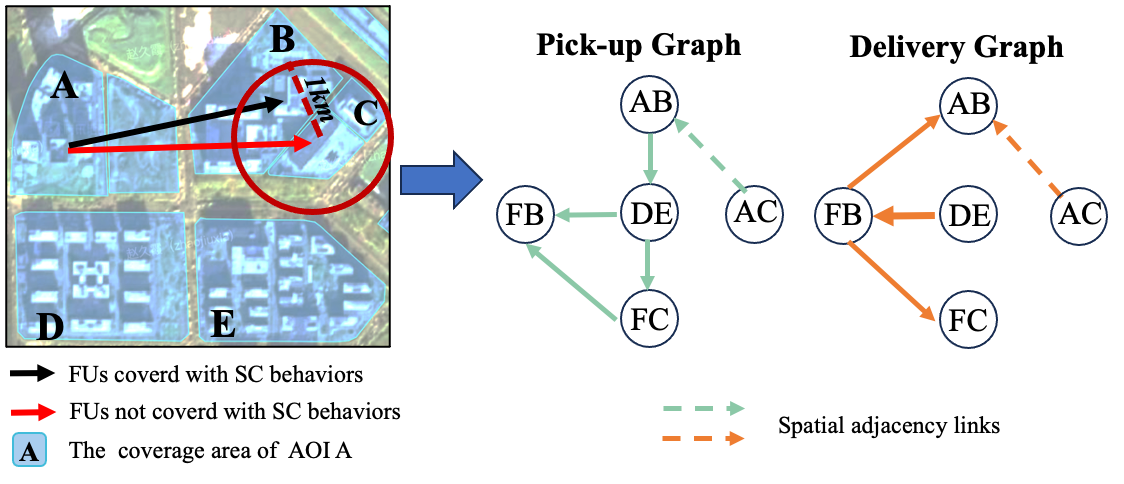}
\caption{Illustration of spatial adjacency relationship in the extended delivery network.} 
\label{fig:spatial_add}
\end{figure}

SC behaviors cover only 60\% FUs. 
To compensate for the loss, we construct an extended delivery network based on geographical adjacency, shown in Figure \ref{fig:spatial_add}.
The criterion for judging spatial adjacency between FUs is the pick-up AOIs should be same \footnote{The emphasis on the same pick-up points is due to existing data analysis and courier feedback.} and the distance between delivery point is less than a threshold (currently 1km). If no adjacent FUs found, we will relax it to only consider the same pick-up AOI as a fallback. 
Then the embeddings of FUs previously unseen, can be estimated by aggregating the embeddings of their existing neighboring FUs in the network constructed above. This increases FU embedding coverage to over 80\%.

\section{Application and Deployment}

\subsection{Model Deployment} As introduced above, FU embeddings are learned from SC behavior data using EATNE. 
Different models are created for diverse scenarios, like weekday/weekend and peak/idle time, due to their significant differences in order structure. 
Moreover, to accelerate training in big cities, we use community detection algorithms to partition the city network into separate regional groups for parallel training at regional group level.

The models are trained using 4 weeks of data across the country. They are trained for less than 2 weeks on 4 NVIDIA Tesla V100 GPUs with 32GB of memory each, and the models get updated every 2 weeks.

\subsection{Information Mining} Leveraging the FU embeddings, we've created a set of indices.

\textbf{(1) High-quality pooling probability} (HPP) quantifies how well multiple orders can be consolidated together, sharing common pick-up and delivery times and travel distances. Since two FUs that consecutively appear in the SC behavior sequence often possess the above traits, this metric is calculated by the cosine similarity between the FU embeddings of these orders, reflecting the frequency of consecutive co-occurrence of the two FUs in SC behavior data. 
\begin{equation}
p_{ij} = cos(\mathbf{v}_i, \mathbf{v}_j), \forall i, j \in V
\end{equation}
Orders with high HPP values can be consolidated and assigned to the same courier to achieve efficient delivery.

\textbf{(2) FU efficiency indicator} (FEI) measures how much an order in this FU improves efficiency, based on how likely it is to be combined with orders from other FUs to form an efficient delivery sequence.
It is calculated by the weighed aggregate of HPPs for the FU and its neighbouring FUs that share same or nearby pick-up or delivery AOIs. The weights are determined by the order volume of those neighboring FUs.
\begin{equation}
\eta_{i} = \sum_{j\in V_i}{p_{ij} \times w_{ij}}, \forall i \in V
\end{equation}
The higher FEI values, the more likely for the order to be efficiently pooled with other orders, thus improving courier efficiency. FEI values are normalized at the city level for ease of comparisons.

\textbf{(3) Scale-effect hotspot} (SEH) for OFD refers to a local network of geographically proximate FUs, wherein the marginal cost and time of delivery for couriers fulfilling orders in this network progressively diminishes, allowing for comprehensive order consolidation within promised delivery time. 
In accordance, FUs in an SEH should have high FEI values, and any pair of FUs in the same SEH exhibit a relatively high HPP. And the total order volume for each SEH should exceed certain criteria.
\begin{equation}
\left.S=\left\{i\in V\middle|\begin{matrix}\eta_i>Thre_\eta;\\p_{ij}>Thre_p,\forall i,j\in S\end{matrix}\right.\right\}
\end{equation}

\subsection{Deployment in Dispatch System} 
\begin{figure}[!ht]
	\centering
	\includegraphics[width=1.0\columnwidth]{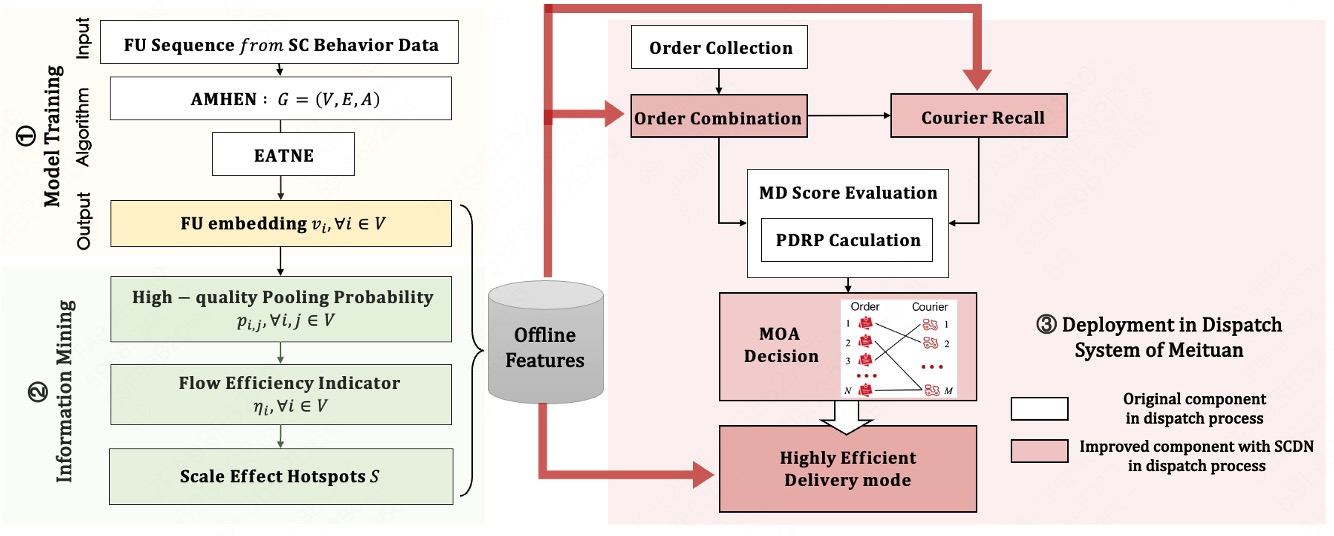}
	\caption{The main execution process of the dispatch system in each dispatch cycle. \label{fig:dispatch_procedure}}
\end{figure}

The above information, including FU embeddings, FEI, SEH, are introduced in the system via offline features. As either low-dimensional vectors or scalars, they are performance-friendly to the real-time storage of the system. The main system framework is shown in Figure \ref{fig:dispatch_procedure}.

\subsubsection{Order Combination and Courier Recall} 

The MOA problem of our system is now solved by well-crafted constructive heuristics, i.e. imitation learning-enhanced iterated matching algorithm (ILIMA) \cite{chen2022imitation}, since metaheuristic algorithms with in-depth search fail to meet the real-time requirements \cite{zhou2020two}. 
Meanwhile, a few orders are combined in mutually exclusive groups based on the closeness of their origins and destinations, as well as promised delivery time, before MD score evaluation.
However, the real-time performance severely restricts the search depth of the algorithm, resulting in insufficient and suboptimal order pooling. 

With SCDN, we develop scalable mechanisms for courier recall and order combination, which can cut down the MOA search space, and let us focus our limited computation time on promising areas. 
Generally, orders with high HPP are formed as favorable combinations in advance, which can greatly expand the proportion of combined orders. Order combinations with low HPP and couriers whose on-hand orders mostly share low HPP with the new order are filtered out. Hence, we can facilitate high-quality order pooling in real time, without obvious increase in score calculation volume and computation time.

\begin{figure}[!ht]
	\centering
	\includegraphics[width=1.0\columnwidth]{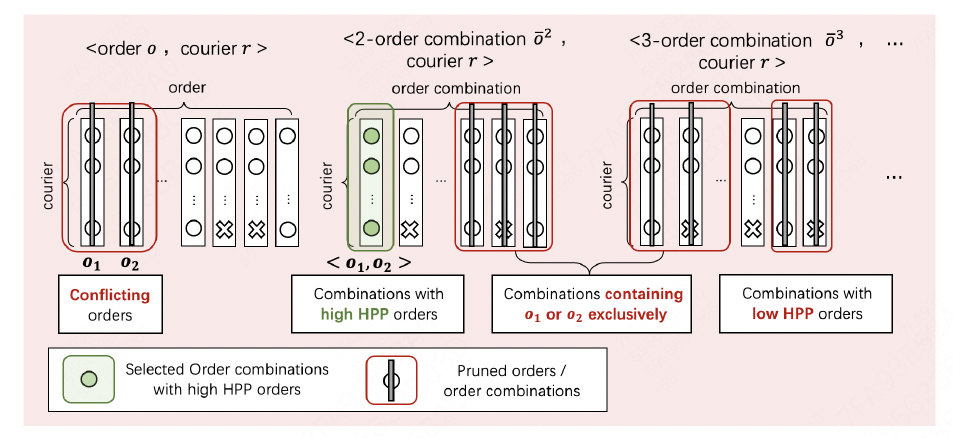}
	\caption{Order combination mechanism pruning MOA search space using HPP information. For example, for candidate orders A, B, C, and combinations AB, AC, BC, if AB and AC have higher HPP, then only AB, AC, B and C are preserved for MD evaluation, while A and BC can be eliminated.\label{fig:combination_details}}
\end{figure}

\textbf{Order Combination.} Based on HPP, high-quality order combinations can be identified and incorporated into ILIMA as expanding decision entities rather than single orders. 
As illustrated in Figure \ref{fig:combination_details}, on one hand, order combinations with very low HPP can be pruned to avoid unnecessary score calculation. 
On the other hand, since top-tier order combinations found by high HPP should be pooled to the same courier, other combinations containing partial orders, and conflicting orders themselves can be removed from the search space.
It can guide ILIMA to search deeply and effectively without obviously increasing score calculation volume.

\begin{figure}[!ht]
	\centering
	\includegraphics[width=1.0\columnwidth]{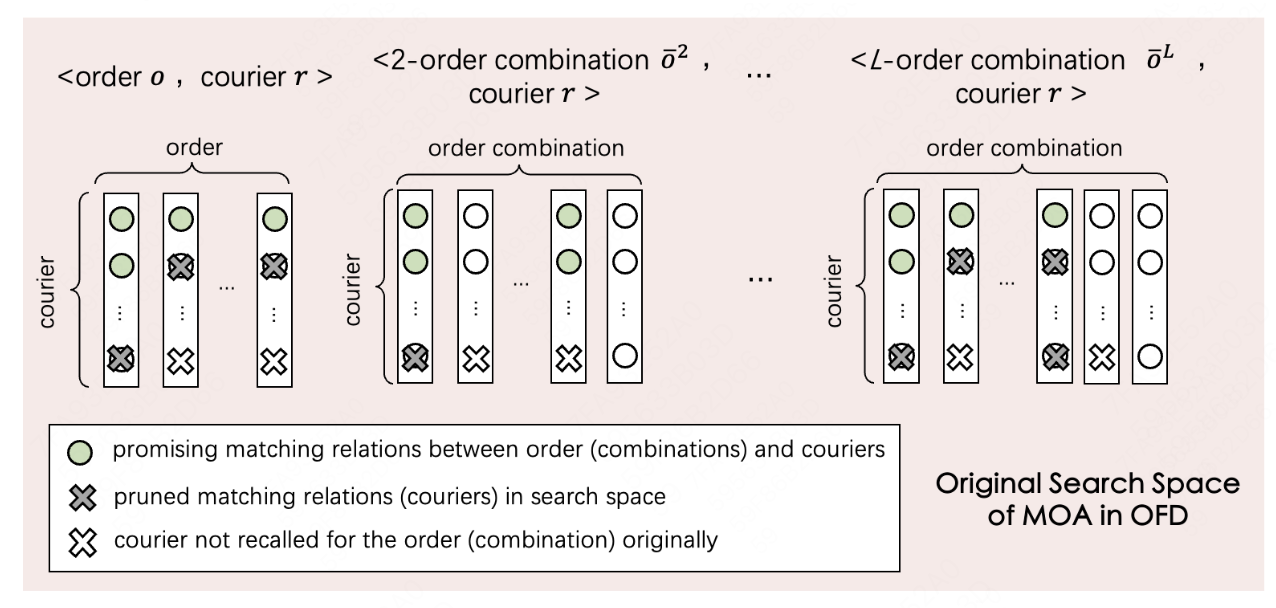}
	\caption{Courier recall mechanisms pruning MOA search space using HPP information. \label{fig:courier_recall}}
\end{figure}

\textbf{Courier Recall.} When retrieving available couriers for an order (combination), we calculate the average value of HPP between it and the courier's on-hand orders, to quickly estimate MD between the order (combination) and the courier, instead of time-consuming score calculations.
For the on-hand orders already picked up by the courier, its FU can be considered as the FU starting from the AOI where the courier is currently located and ending at its delivery AOI.
This further helps to prune the MOA search space and reduce real-time computational pressure while maintaining solution quality, as shown in Figure \ref{fig:courier_recall}. 

The implementation details of order combination and courier recall mechanisms can be found in Appendix \ref{implementation_details}.

\subsubsection{Highly Efficient Delivery Mode} 
\begin{figure}[!ht]
	\centering
	\includegraphics[width=1.0\columnwidth]{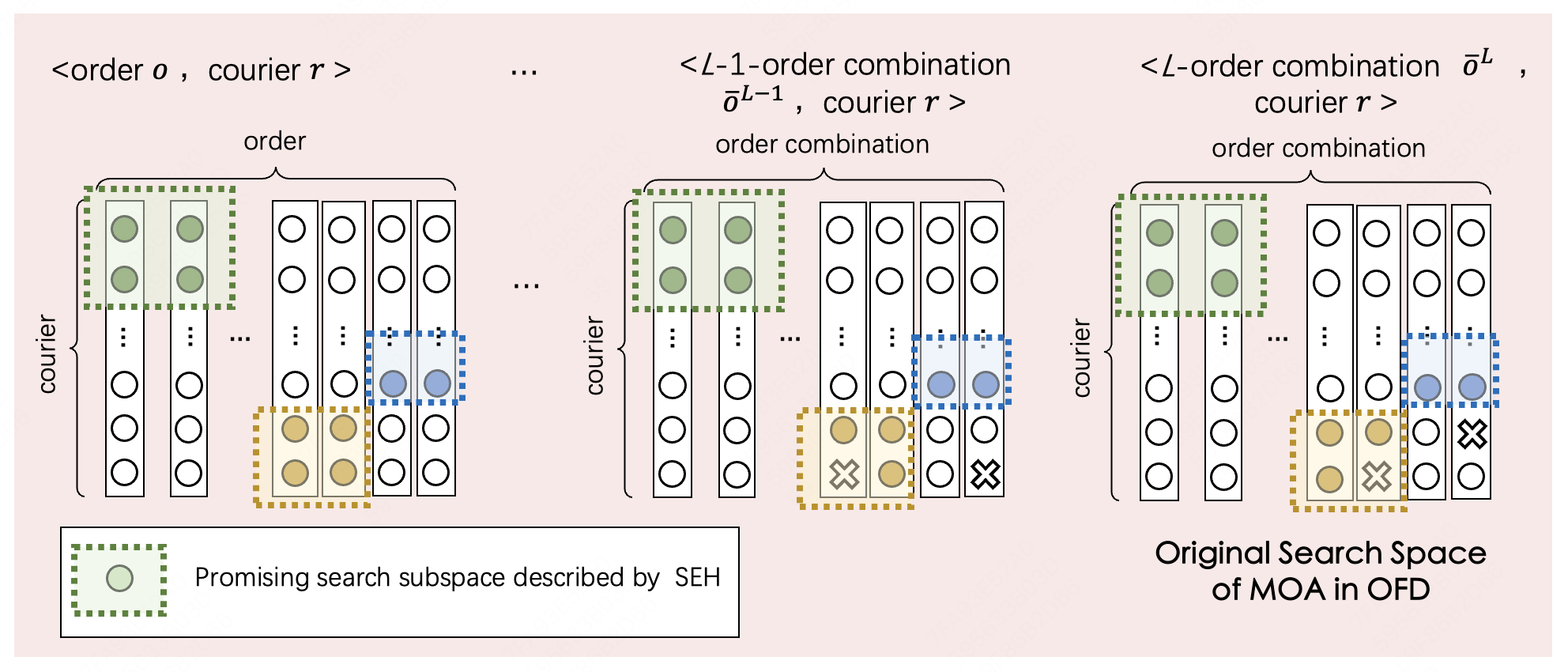}
	\caption{Promising MOA search subspace described by SEH. \label{fig:subspace_seh}}
\end{figure}

SEHs identified by SCDN, essentially represent small-scale subspaces deeply embedded within the MOA search space, where thorough and high-quality order pooling outcomes can be found, as shown in Figure \ref{fig:subspace_seh}.
Then a new delivery mode can be built, wherein a dedicated group of couriers is assigned to each SEH, as opposed to receiving assignments in the entire region. Accordingly, the original large-scale MOA problem, initially solved within a vast search space shown in Figure \ref{fig:challenge_OFD_1}, can be effectively decomposed into \textbf{a collection of small-scale MOA problems, defined within much smaller and independent subspaces, paving the way for comprehensive and in-depth real-time searching}. This approach serves to continually enhance the courier efficiency potential.

In the delivery mode, order assignments for each SEH are conducted as follows:

\textbf{(1) Hourly SEH Identifications.}
SEHs for certain time periods in a city are found using binary programming (BP), which categorizes FUs with high FEI within a specified time period into a number of mutually exclusive sets. It aims to maximize the average HPP among FUs within each set, with FU quantities and total historical order volume in each set as constraints. 
Practically, in some mega cities like Beijing, SEHs in peak periods are determined every 30 minutes to capture the changes in order structure. The BP problems for SEH identification can be solved via genetic algorithm \cite{mirjalili2019genetic} within 10 minutes.
More information is in Appendix \ref{APP-SEH}.

\textbf{(2) Real-time Parallel MOA Solutions.}
Order assignment for SEH is a scaled-down MOA problem. 
Given the limited area and stable order structure for SEH in a certain time period, the behavioral patterns of mode couriers are highly certain, thus simplifying the MD evaluation. 
In reality, we evaluate the MD via a weighted sum of average order increments for pick-up and delivery AOIs in a courier's route after the new order acceptance for SEH, instead of time-intensive PDRP calculations to simulate couriers' routes. 
Hence we can evaluate the MD between any promising order combinations and candidate couriers in real time, and solve the completely-modeled MOA problem for each SEH using a HillClimbing heuristic algorithm \cite{zhang2017taxi} in parallel, helping to pool orders effectively and thoroughly.
Orders outside SEHs keep the existing assignment rules.

\section{Experimental Evaluation}
\subsection{Model Performance Evaluation}
\subsubsection{Model Learning Performance}
Link prediction task is used to evaluate the performance of EATNE, with AUC, F1 score and PR as evaluation criteria.
The experiments are conducted on a real-world dataset collected from Meituan delivery platform, using a
single Linux server with NVIDIA Tesla V100 GPU with 32GB memory.
The dataset contains 28,000 SC behavior records from 28 days in Beijing, China, forming a delivery network with about 70,000 FUs. 
For each edge type, the test set is generated with 10\% randomly chosen positive edges and an equal number of negative edges, selected by regional negative sampling. Parameter details are in Appendix \ref{APP-model}.

\begin{figure}[!ht]
\centering
\includegraphics[width=0.7\columnwidth]{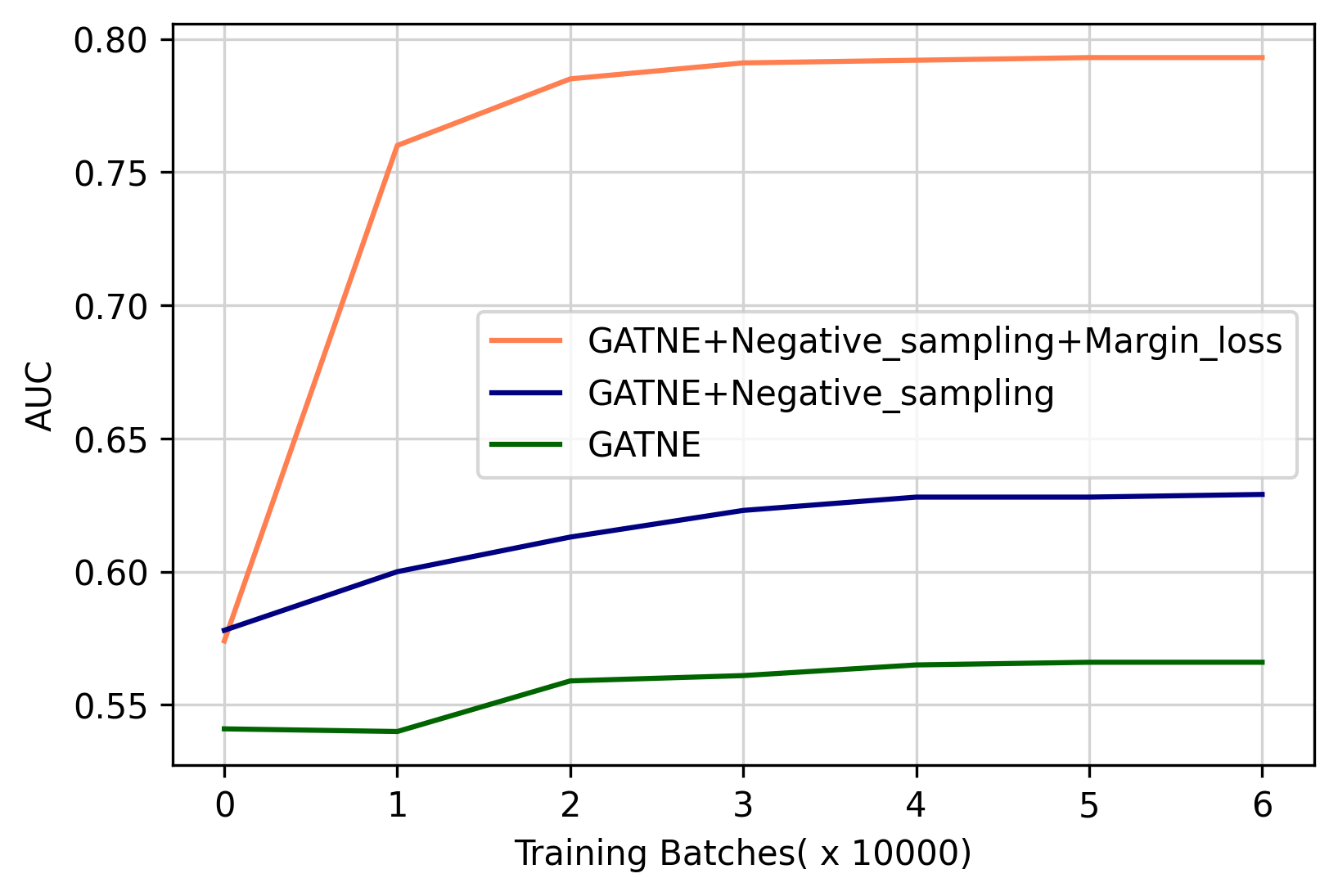}
\caption{The convergence curve for different algorithms.}
\label{fig:margin-loss}
\end{figure}

First we examine the effectiveness of EATNE. Figure \ref{fig:margin-loss} shows that the original GATNE is hard to converge in this situation. While EATNE, armed with regional negative sampling and margin loss, produces superior outcomes in addition to converging much faster.
Next the performance of EATNE in various graph configurations is investigated.
Table \ref{tab:model_evaluate} shows that optimal performance is achieved by graphs with pick-up and delivery edges and node attributes, proving the validity of the proposed ANHEN.
Notably, pick-up connections are more important than delivery ones, indicating pick-up behaviours have a greater effect on courier efficiency. Moreover, adding node attributes is highly impactful, highlighting order structure's key role in affecting courier efficiency.
\begin{table}[h]
\centering
\caption{Model performance under different graph settings}
\label{tab:model_evaluate}
\begin{tabular}{p{0.12\linewidth}p{0.12\linewidth}p{0.12\linewidth}p{0.12\linewidth}p{0.12\linewidth}p{0.12\linewidth}}
\hline
Node Attr. & Pick-up Edge & Delivery Edge & AUC & F1 & PR \\
\hline
\checkmark & \checkmark & \checkmark & \textbf{0.79} & \textbf{0.72} & \textbf{0.75} \\
\ding{55} & \checkmark & \checkmark & 0.64 & 0.60 & 0.59 \\
\checkmark & \ding{55} & \checkmark & 0.74 & 0.69 & 0.71 \\
\checkmark & \checkmark & \ding{55} & 0.76 & 0.71 & 0.73 \\
\hline
\end{tabular}
\end{table}

\subsubsection{FU Embedding Effectiveness}
\begin{figure}[!ht]
\centering
\includegraphics[width=0.9\columnwidth]{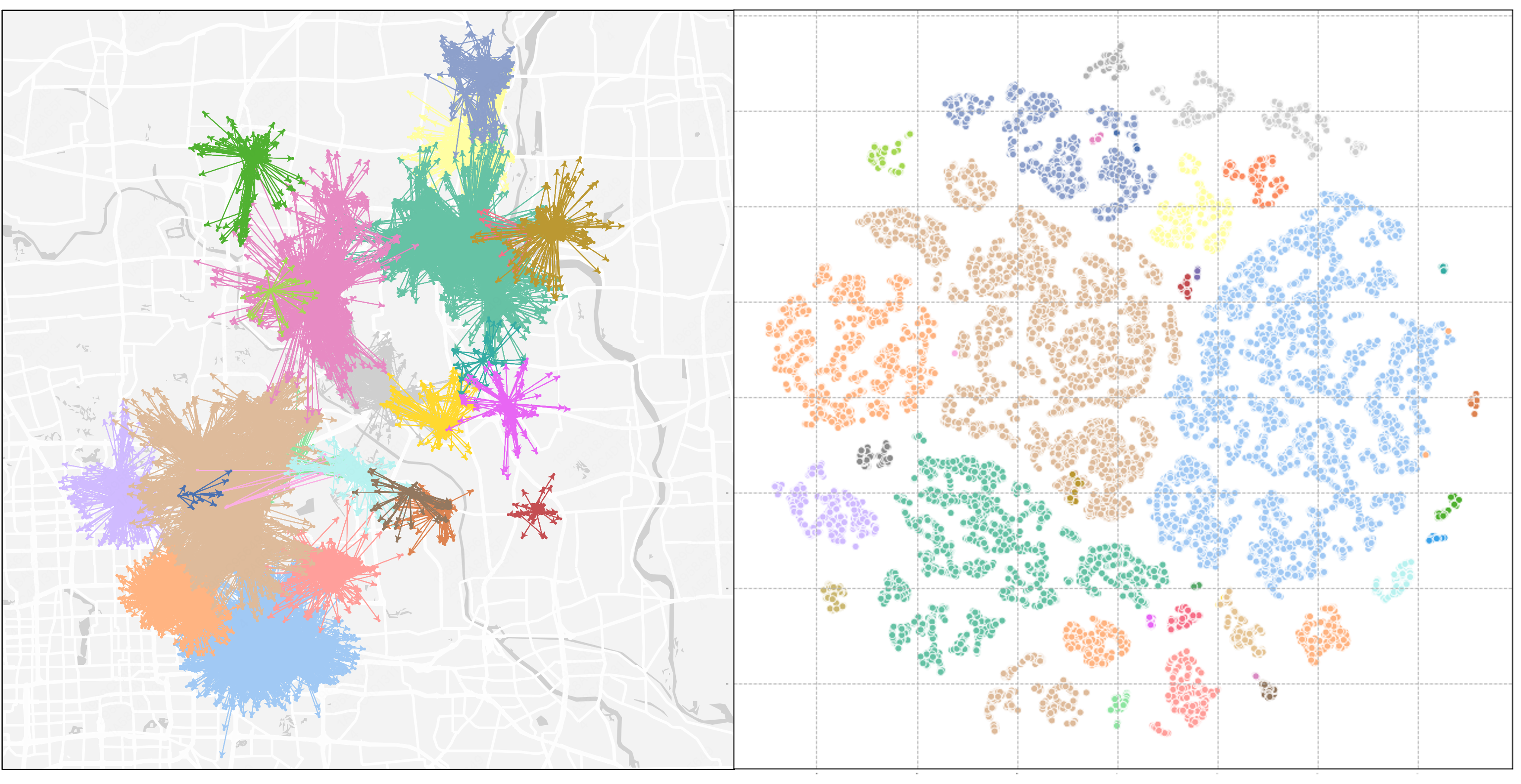}
\caption{FU embedding clusters of a district in Beijing on map (left) and after T-SNE (right).}
\label{fig:flow-embedding}
\end{figure}

To evaluate the effectiveness of FU embeddings, we examine the training results via the data of the same district in Beijing. First, by performing DBSCAN clustering on learned embeddings, we evaluate if geographical similarity is encoded. Figure \ref{fig:flow-embedding}, which shows resulting 33 clusters, confirms the FUs from close locations are clustered together in the hidden space. 

\begin{figure}[!ht]
  \centering
  \subfigure[FU pairs with high similarity.]{
    \label{fig:flow-similarity:a}
    \includegraphics[width=0.9\columnwidth]{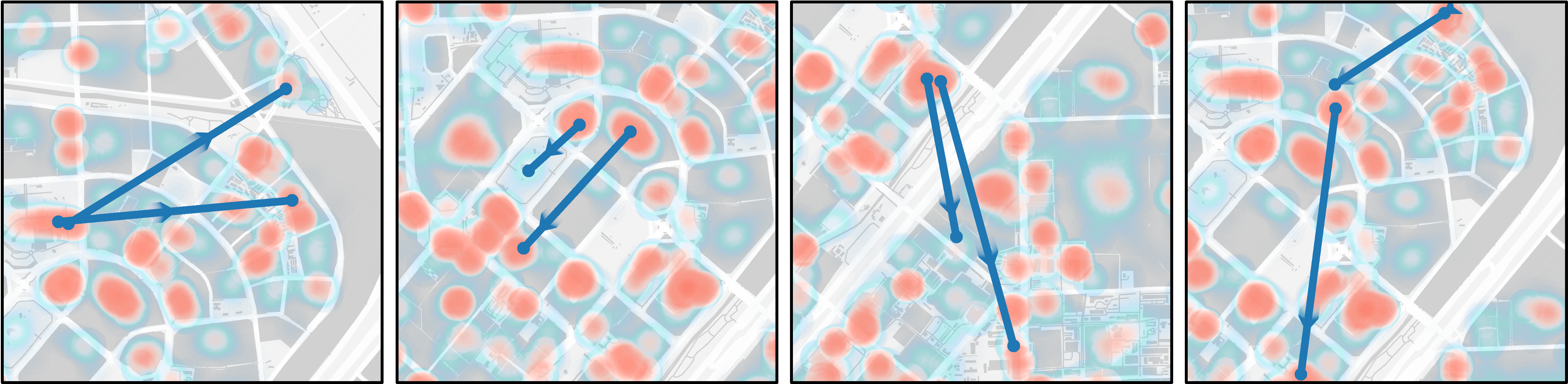}
  }
  \subfigure[FU pairs with low similarity.]{
    \label{fig:flow-similarity:b}
    \includegraphics[width=0.9\columnwidth]{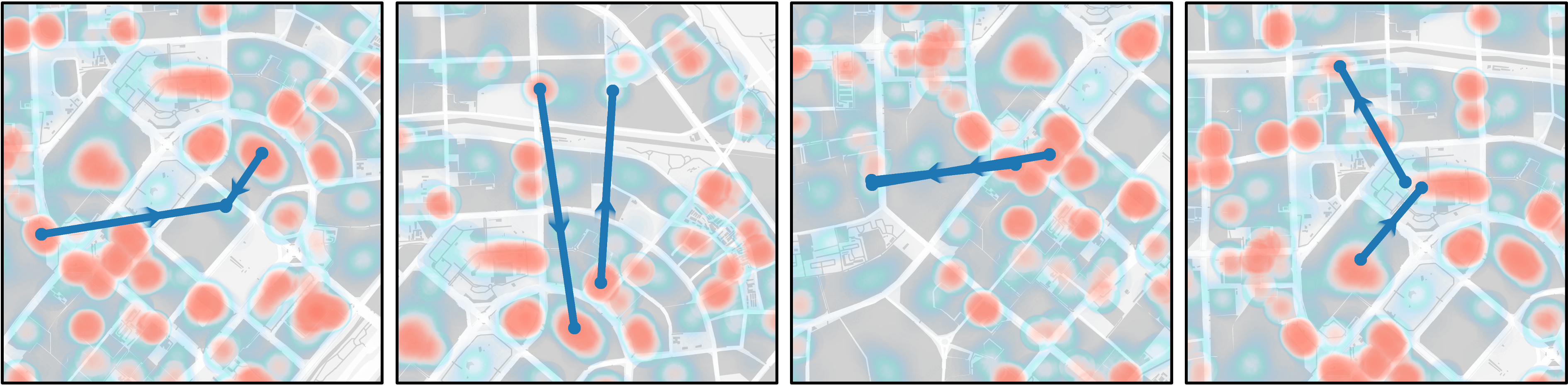}
  }
  \caption{FU pair cases in different similarity levels.}
  \label{fig:flow-similarity}
\end{figure}

Next we demonstrate high-quality pooling potential can be captured by FU embedding similarity, i.e. HPP. Figure \ref{fig:flow-similarity:a} shows four cases of FU pairs with high HPP, including (1) FU pair with pick-up and delivery AOIs located closely, (2) nearby parallel FU pair, (3) FU pair where one runs alongside the other, and (4) head-to-tail FU pair, with the tail one pointing high-order-density AOIs, leading to less courier empty run time \footnote{Empty run time refers to the empty cruising time before carriers deliver next orders.} after completing deliveries. Orders in these FU pairs can be pooled for simultaneous delivery to improve courier efficiency.
Meanwhile, we also identify FU pairs with low HPP. Figure \ref{fig:flow-similarity:b} illustrates four cases of this situation, including (1) FU pair with the same delivery AOI but pick-up AOIs located far apart, (2) reverse parallel FU pair, (3) FU pair where one FU runs alongside the other but points a low-order-density AOI, leading to longer courier empty run time, and (4) head-to-tail FU pair that also leads to a low-order-density area. These FU pairs are unlikely to be efficiently pooled together and may undermine courier efficiency.
\subsection{Order Combination and Courier Recall}

The proposed method, ILIMA + SCDN, is evaluated against the current online implementation, which utilizes ILIMA with ruled batching method, and MNDS, a metaheuristic algorithm used in \cite{chen2022imitation}. 
Experiments are conducted in a mid-sized Chinese city, involving around 500 orders and 2,500 couriers in a dispatch cycle during noon peak.

\begin{figure}[!ht]
  \centering
  \subfigure[ILIMA+Rule.]{
    \label{fig:combination_level:base}
    \includegraphics[width=0.3\columnwidth]{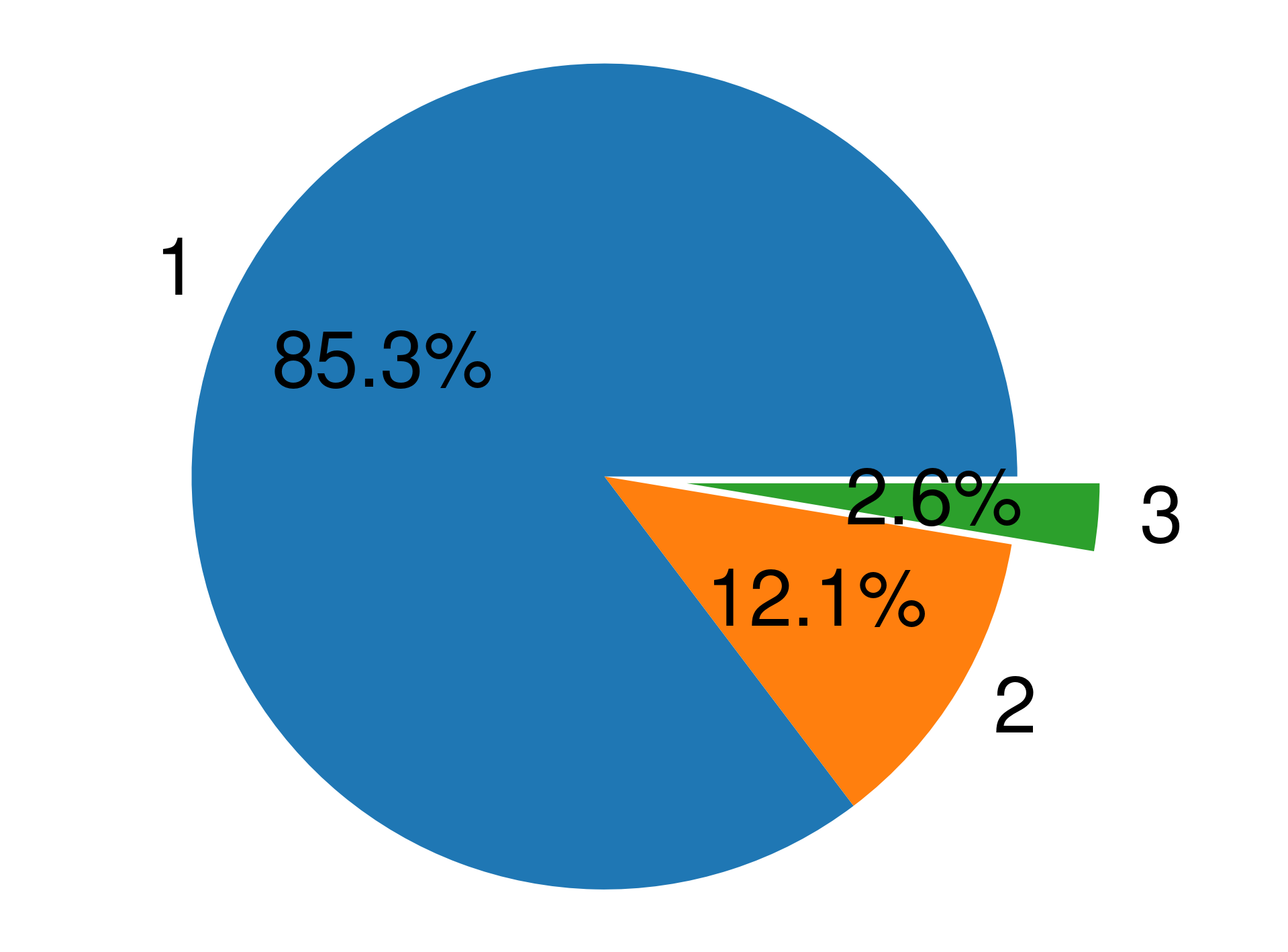}
  }
   \subfigure[ILIMA+SCDN.]{
    \label{fig:combination_level:scdn}
    \includegraphics[width=0.3\columnwidth]{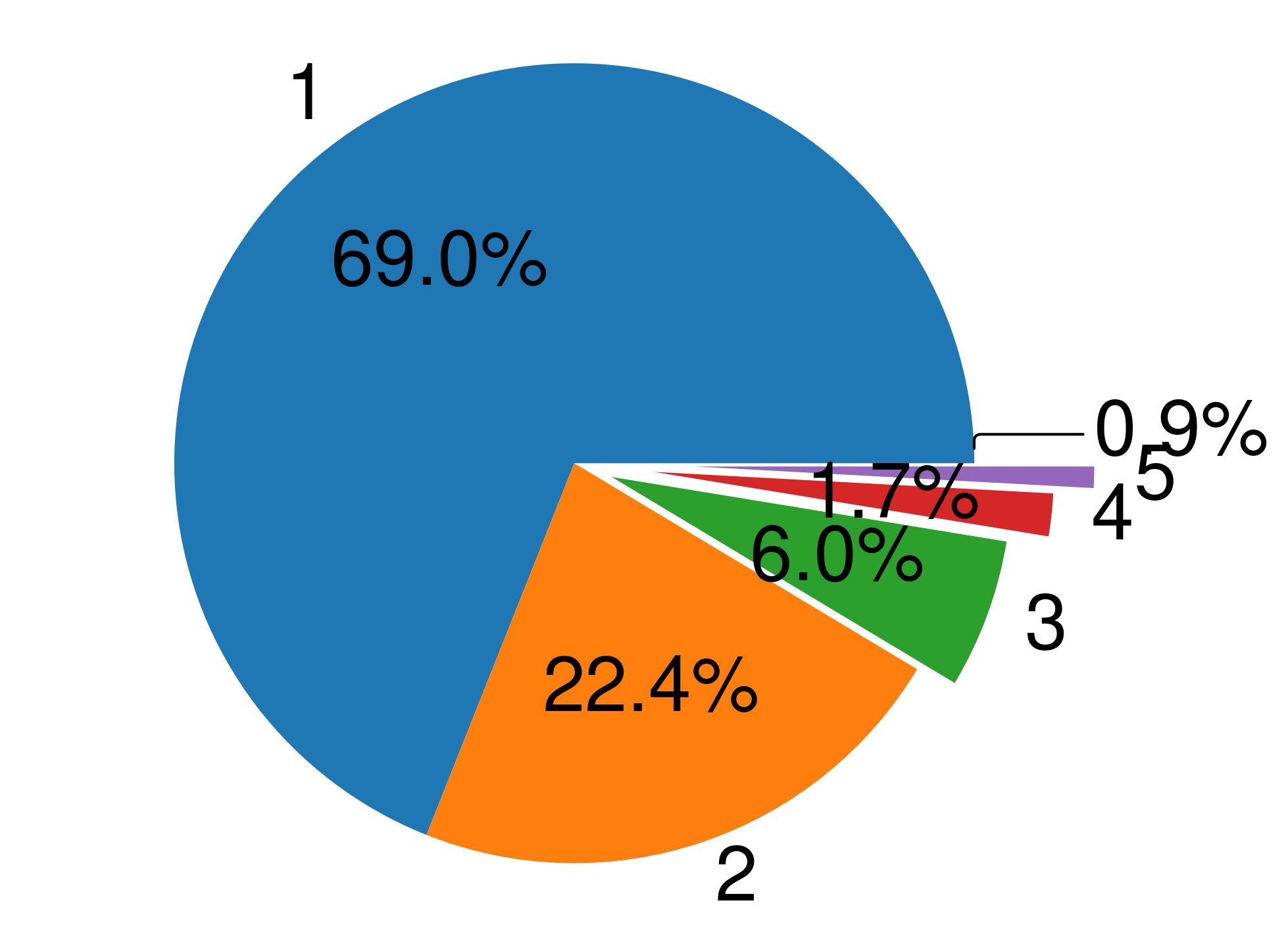}
  }
  \subfigure[MNDS.]{
    \label{fig:combination_level:mnds}
    \includegraphics[width=0.3\columnwidth]{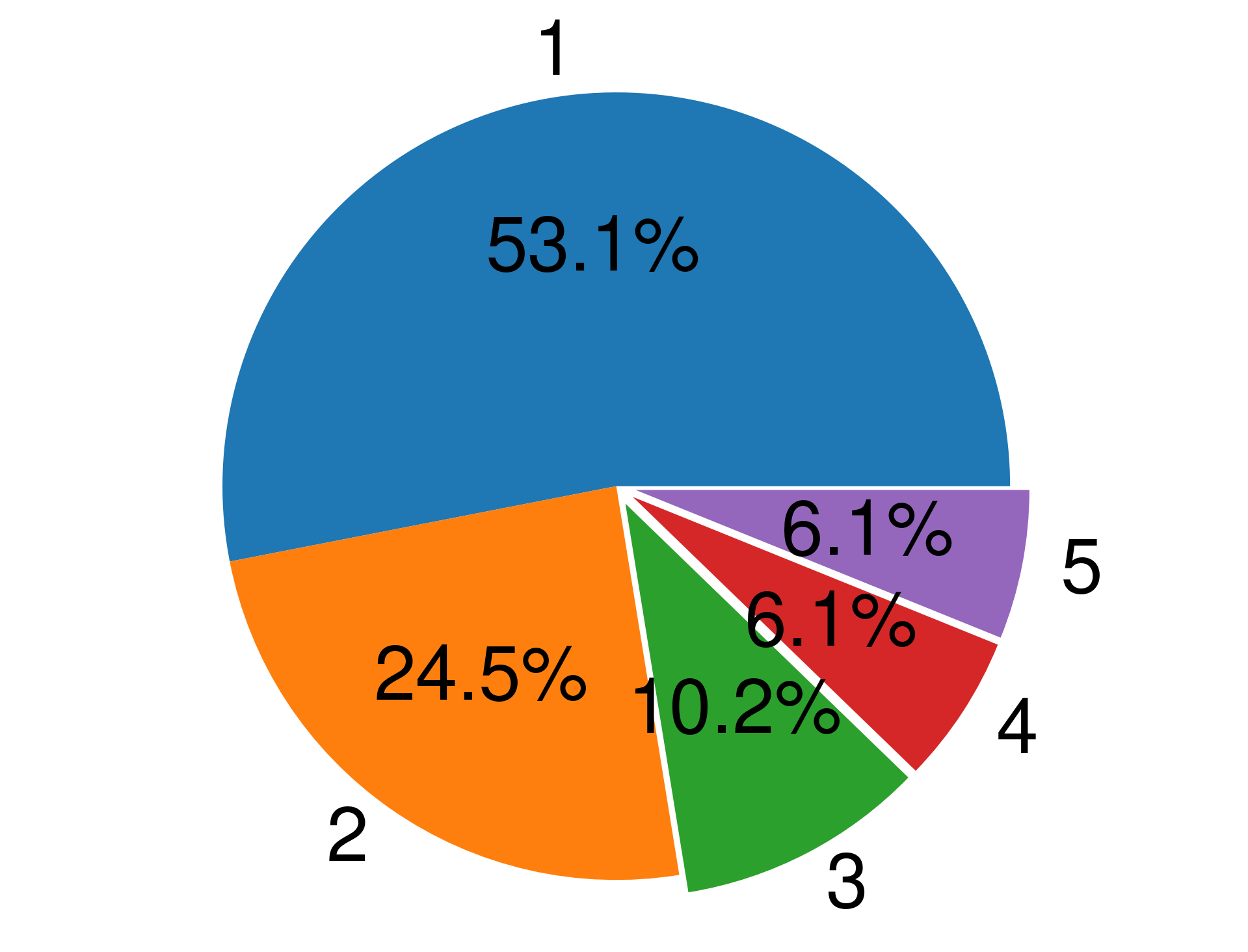}
  }
  \caption{Combination level distribution.}
  \label{fig:combination_level}
\end{figure}

The comparison results on both computational cost and solution quality are presented in Table \ref{tab:compare_moa}. 
The ILIMA+SCDN approach enhances the total MD score of MOA solutions by 5.3\% compared to ILIMA+Rule method, without incurring a significant increase in time consumption. 
However, it lags by 1.2 $pp$ behind MNDS. Despite this, MNDS requires exploration of a much larger search space and massive PDRP calculations, which takes over 20 seconds on average, making it unsuitable for online use.
Hence, the proposed method excels at balancing computational time and solution quality, securing more optimal MOA solutions in real-time.
Moreover, Figure \ref{fig:combination_level} illustrates that the overall combination level grows as the percentage of couriers assigned only one order decreases by 16.3 $pp$.
This shift results in increasing order consolidation.
Online A/B test show that while maitaining delivery experience, couirer efficiency, i.e.orders completed per hour, is augmented by 3.7\%.

\begin{table}[!ht]
\centering
\caption{Computation cost and score improvement of MOA.}
\label{tab:compare_moa}
\begin{tabular}{cccc}
\toprule
Method& \multirow{2}{*}{\shortstack{Online PDRP\\ Calculations}}  & \multirow{2}{*}{\shortstack{ Computation \\ Time Online/s}}  &  \multirow{2}{*}{\shortstack{MD Score \\ Improvement}}    \\
& & & \\ \midrule
ILIMA+Ruled & 44,541  & 5.6 & 0\% \\
ILIMA+SCDN & 48,998  & 6.9 & 5.3\%\\ 
MNDS & /  & / & 6.5\%\\ 
\bottomrule
\end{tabular}
\end{table}

Table \ref{tab:various_order} presents the results of offline experiments conducted with varying order volumes. In different order size scenarios, the proposed ILIMA+SCDN method significantly enhances the MD score over the existing ILIMA+Ruled method.
Regarding PDRP Calculations, for orders fewer than 400, our proposed ILIMA+SCDN method demonstrates lower PDRP Calculations compared to the ILIMA+Ruled method. Nevertheless, as the order volume escalates, the computational burden of both methods exhibits nearly linear growth, aligning with the online time requirements.

\begin{table}[!ht]
\centering
\caption{MOA results across various order sizes.}
\small
\setlength{\tabcolsep}{3.5pt}  %
\label{tab:various_order}

\begin{tabular}{lccccc}
\toprule
Method & (0, 200] & (200, 400] & (400, 600] & (600, 800] & (800, 1000] \\
\midrule
\multicolumn{6}{l}{MD Score Improvement} \\
ILIMA+Ruled & 0\% & 0\% & 0\% & 0\% & 0\% \\
ILIMA+SCDN & 1.0\% & 4.0\% & 4.4\% & 5.5\% & 3.7\% \\
MNDS & 1.7\% & 5.3\% & 5.3\% & 6.9\% & 5.6\% \\
\midrule
\multicolumn{6}{l}{PDRP Calculations} \\
ILIMA+Ruled & 4,285 & 21,910 & 37,323 & 57,700 & 79,596 \\
ILIMA+SCDN & 4,250 & 20,589 & 38,358 & 65,011 & 94,292 \\
\bottomrule
\end{tabular}
\end{table}

\subsection{Highly Efficient Delivery Mode}
\begin{figure}[!ht]
	\centering
	\includegraphics[width=0.9\columnwidth]{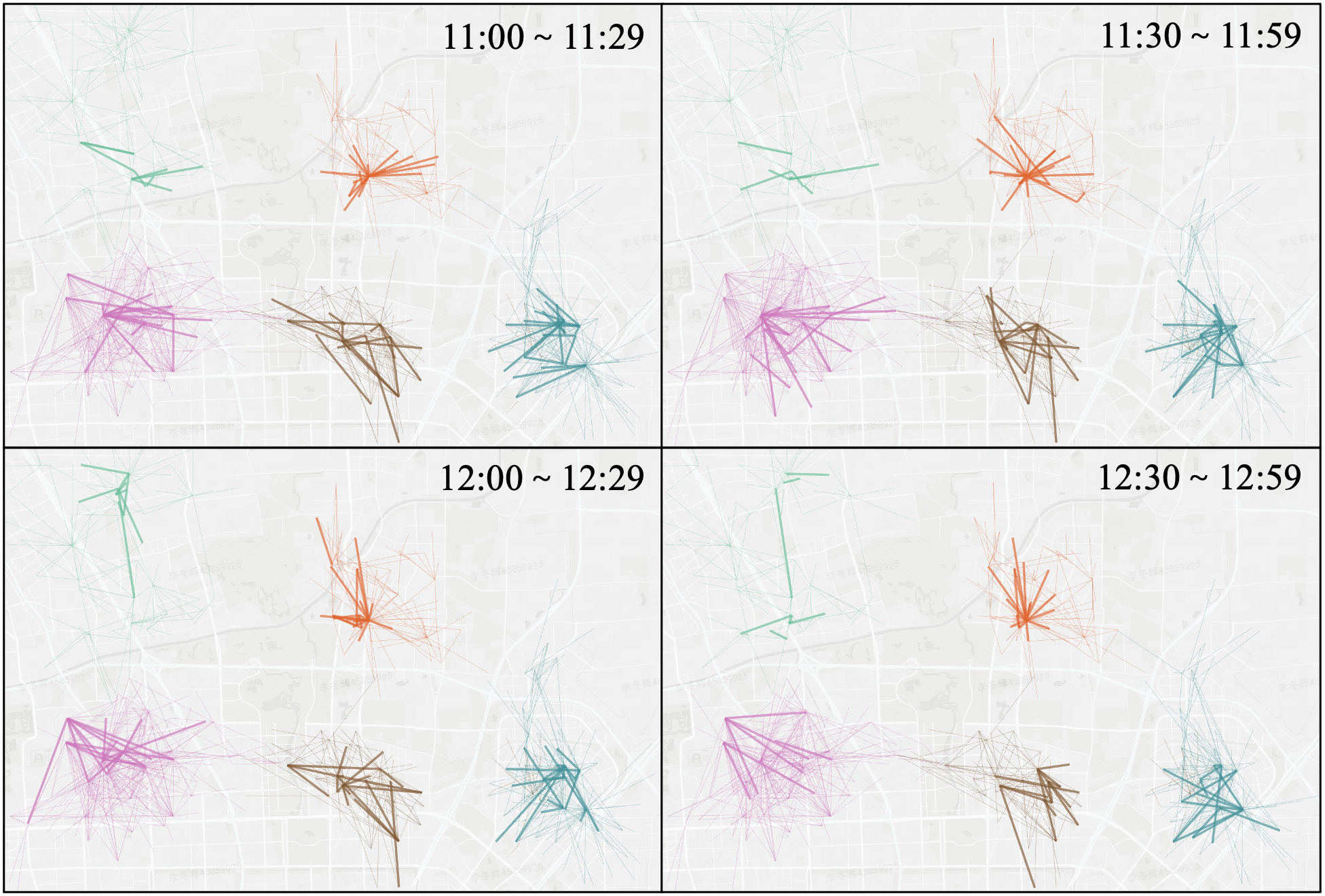}
	\caption{SEHs over time, with each image capturing half an hour. Colors denote different areas, bold lines for internal SEH FUs, and thin lines for external FUs. \label{fig:clusters}}
\end{figure}

Figure \ref{fig:clusters} depicts 5 SEHs identified in a specific district of Beijing during weekday noon peak period (11:00-12:59).  In response to fluctuations in order structures, the network configuration of each SEH is updated every half hour. 
On average, each SEH processes about 81 orders every half hour with an average HPP of 0.65, ensuring high order density and strong network connectivity. 
Moreover, the maximum number of orders pending assignment in each cycle is less than 10. By allocating 5 to 8 couriers per SEH, we significantly simplify the complexity of MOA solutions for each SEH.

\begin{figure}[!ht]
	\centering
	\includegraphics[width=1.\columnwidth]{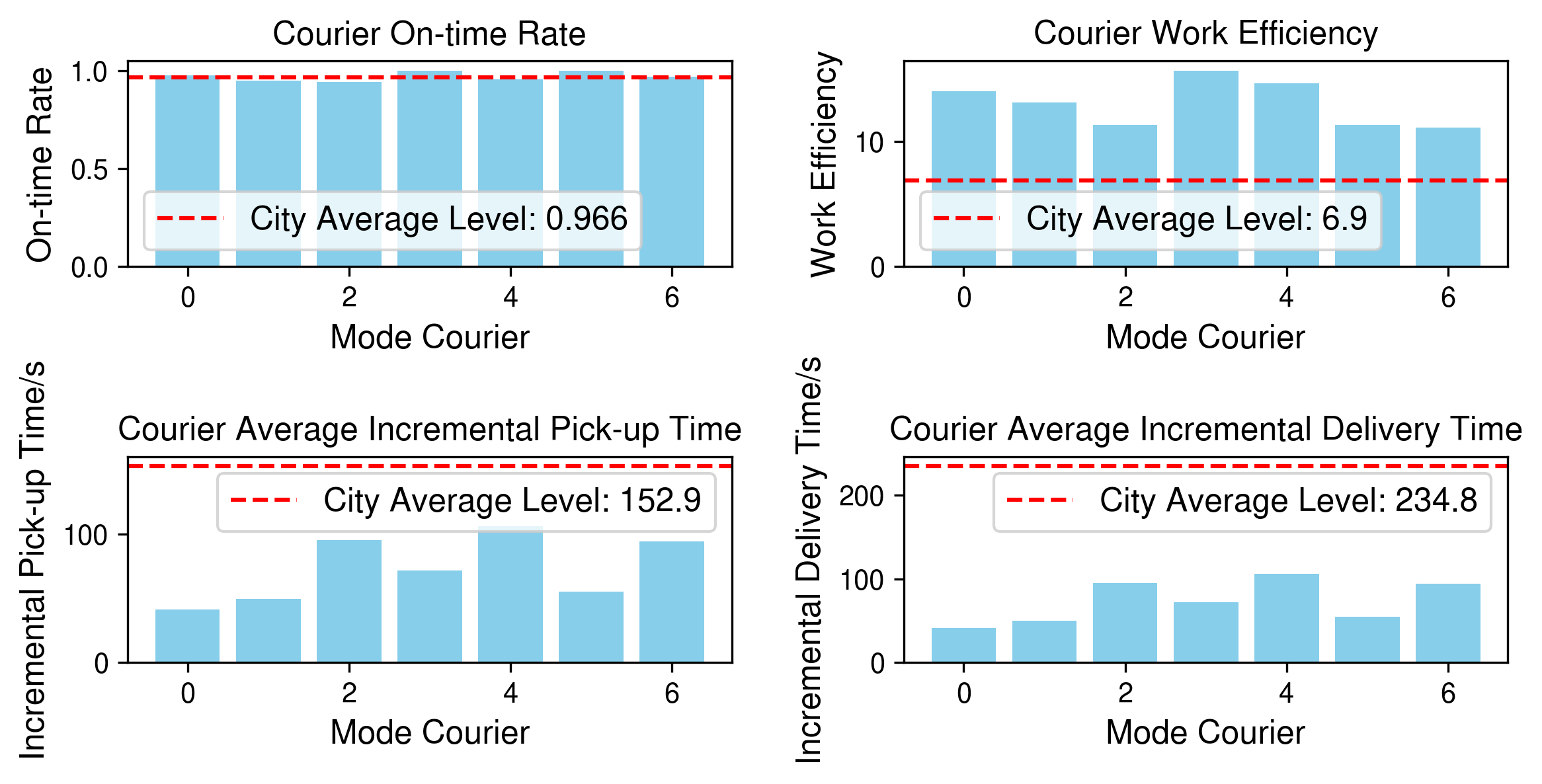}
	\caption{Courier performance in a SEH mode in noon peak. \label{fig:modes}}
\end{figure}

Taking a SEH in Beijing as an example, online tests show a major boost in order pooling. During noon peak, a courier can accept over 7 orders at once. And the percentage of SEH couriers picking up over 5 orders simultaneously in the same AOI has risen by 23.5 \textit{pp} compared to past performance. Likewise, the percentage of SEH couriers delivering over 5 orders at once in the same AOI has increased by 20 \textit{pp}.
The average courier incremental pick-up time has been reduced by 51\% and delivery time by 21\%.
These enhancements lead to a 45-55\% boost in courier efficiency, i.e.orders completed per hour, while maintaining consistent work hours and on-time delivery standards. Figure \ref{fig:modes} illustrates the superior performance of SEH mode against city average level in noon peak, where each bar corresponds to the trial performance of a specific courier.
\section{Conclusion}
This paper proposed a systemic solution framework, SCDN, based on an Enhanced GATNE method tailored for OFD, to resolve real-time OFD order pooling problem. It uncovers the latent potential for order pooling embedded within SC trajectories, which can strengthen system awareness and effectively inform decisions. 
Accordingly, the vast search space of NP-hard MOA problems in OFD is effectively pruned through scalable similarity calculations of simple vectors. Thus high-quality and comprehensive pooling outcomes are found in real time.
Moreover, the outcomes highlight SEHs for OFD, where highly-efficient delivery modes are built for continuously improving efficiency. SCDN has now been deployed in Meituan. Online tests show it has achieved excellent performance and well-acknowledged by all the stakeholders.

\bibliographystyle{ACM-Reference-Format}
\balance
\bibliography{sample-base}

\appendix
\section{Many-to-one Assignment Problem at Each Dispatch Cycle}\label{APP-MOA}

As shown in Figure \ref{fig:challenge_OFD_1}, the calculation volume increases very fast with the number of orders and couriers. Different order combinations of order set $O_t$ are considered. For example, the number of $l-$order combinations is $C_{|O_t|}^l$. Since the MD score of assigning combinations of orders is not equivalent to the sum of scores of individual assignments. The calculation volume of MD score is $\sum_{l \in L} C_{|O_t|}^l \times |\cup_{\forall \bar{o}^l} R_t^{\bar{o}^l}|$, where $R_t^{\bar{o}^l}$ is the set of couriers for $l-$order combination ${\bar{o}^l}$ at dispatch time $t$.

\begin{equation}
\begin{gathered}
\min_{ x_t \in \aleph_t } \sum_{ \bar{o}\in comb (O_t)}\sum_{r\in R_t^{\overline{o}}}\left(\sum_{g\in G} \eta_t^{g} \times f_{t,r}^{g,\bar{o}} \right) \times x_r^{\bar{o}} \\
\left.s.t.\aleph_{t}=\left\{\begin{aligned}&\sum_{\overline{o}(o)}\sum_{r\in{R}_{t}^{\overline{o}(o)}}x_{r}^{\bar{o}(o)}=1,\forall o\in\mathcal{O}_{t}\\&x_{r}^{\bar{o}}=\prod_{o\in\bar{o}}x_{r}^{o},\forall\bar{o}\in comb(O_{t})\end{aligned}\right.\right\} 
\end{gathered}
\label{eq:moa}
\end{equation}

After getting all these MD scores, the MOA problem can be formulated into an integer programming problem in Equation~(\ref{eq:moa}). The objective function is to minimize the total MD scores for different goals, and $f_{t,r}^{g,\bar{o}}$ is the MD score of assigning order combination $\bar{o}$ to courier $r$ at time $t$ for goal $g$, $comb(O_t)$ refers to all the possible combinations constructed by orders in $O_t$, $\eta_t^g$ is the weight of goal $g$ in the objective function at time $t$. The constraint is to make sure each combination $\bar{o}$ can only be assigned to one courier and only one combination of each order can be selected. $\bar{o}(o)$ represents the order combination containing order $o$.

\section{Definition of Skilled Courier and Selection Criteria of Route Sessions}\label{APP-sc}
As mentioned above, SC refers to the couriers with relatively high efficiency, currently set top rank 5\%-35\% in a delivery region.
It should be noted that in order to prevent extreme cases from affecting the validity of the learning outcomes, the top 5\% of couriers have been excluded. 

The SC route sessions of both pick-up and delivery type, for constructing the network are selected based on the following criteria:

(1) time interval between the execution of two consecutive orders less than 30 minutes; 

(2) no overtime orders;

(3) no speeding behaviours; 

(4) no orders with negative feedback reported. 

Then based on the carefully selected sessions of SCs, we construct the corresponding AMHEN using the method outlined in Section \ref{method}.

\section{FU Node Attributes in AMHEN}\label{APP-node}
We incorporate rich spatial and temporal information as attributes of a FU node, for a specific scenario (i.e., weekday/weekend, peak/idle time), mainly including: 

(1) average order volume of FU, and the corresponding pick-up and delivery AOIs in the scenario for last 30 days;

(2) average meal-waiting and pick-up time duration of the corresponding pick-up AOI in the scenario for last 30 days ;

(3) average delivery time duration of the corresponding delivery AOI in the scenario for last 30 days ;

(4) average delivery distance of the FU;

(5) average FU delivery period of time since consumers order in the scenario for last 30 days; 

(6) type and number of natural barriers (e.g. bridge, river, high-way) along the FU path; 

(7) latitudes and longitudes of the center points of the corresponding pick-up and delivery AOIs; 

(8) the proportion of SCs who chose the corresponding pick-up and delivery AOIs as their preferred locations for the scenario in the past 30 days.

\section{Implementation of EATNE Algorithm}\label{APP-eatne}
The proposed EATNE algorithm is summarized in Algorithm \ref{alg:algorithm}.
\begin{algorithm}[!ht]
\caption{EATNE for OFD}
\label{alg:algorithm}
\textbf{Input}: Network $G$; 
  Embedding dimension $d$; 
  Edge embedding dimension $s$;
  Window size $c$; 
  Learning rate $\eta$; 
  Marigin loss min distance $m_p$, $m_d$;
  coefficient $\alpha, \beta, \gamma_p^P, \gamma_d^P, \gamma_p^D, \gamma_d^D$ .
  \\
\textbf{Output}: Embedding $\mathbf{v_i}$, and Embeddding $\mathbf{v}_{i, p}$ and $\mathbf{v}_{i, d}$ on the pick-up and delivery edge for all $v_i \in V$.
\begin{algorithmic}[1] %
\State Initialize all the model parameters $\theta$.
\State Generate positive data sets $\mathcal{D}_p^P$ and $\mathcal{D}_d^P$ by random walk on the pick-up and delivery edge, respectively.
\State Randomly sample FU pairs within the same delivery region, then add to negative data set $\mathcal{D}^N_p$ and $\mathcal{D}^N_d$.
\While{\textit{not converged}}
  \For{each FU pair in ${\mathcal{D}_p^P,\mathcal{D}_d^P}$}
    \State Calculate $\mathbf{v}_{i, p}$ and $\mathbf{v}_{i, d}$ using Equation (4) and (5) respectively;
    \State Sample $m$ negative samples and calculate loss value using Equation (6).
    \State Update model parameters $\theta$ by $\frac{\partial E}{\partial \theta}$.
  \EndFor
\EndWhile
\State  Set $\mathbf{v_i}$ as the average of $\mathbf{v}_{i,p}$ and $\mathbf{v}_{i,d}$.
\end{algorithmic}
\end{algorithm}

\section{EATNE Model Parameter Configuration}\label{APP-model}
The detailed parameter setting is shown in Table \ref{tab:parameter}. We employ the Adam optimizer with default settings for training. The model implements early stopping if there's no improvement in the ROC-AUC on the validation set within a single training epoch.
\begin{table}[!ht]
\centering
\caption{Parameter configuration of EATNE model.}
\label{tab:parameter}
\begin{tabular}{ccc}
\toprule
Notation & Description&  Setting Value       \\ \midrule
$d$ & base embedding dimension  & 200 \\
$s$ & edge embedding dimension  & 20\\ 
$l$ & random walk length  & 10\\ 
$c$ & sampling window size  & 3\\ 
$m_p$, $m_d$ & margin loss min distance  & 0.3\\ 
$\eta$ & learning rate  & 0.001 \\
$\alpha_p$, $\alpha_d$,$\beta_p$,$\beta_q$& edge weights & 1 \\
$\gamma_p^P, \gamma_d^P, \gamma_p^D, \gamma_d^D$& weights in loss objective & 1 \\
\bottomrule
\end{tabular}
\end{table}

\section{Implementation Details of Order Combination and Courier Recall.}\label{implementation_details}
The MOA problem in our system is now solved using a constructive heuristic framework. The process during each dispatch cycle may require multiple iterations. Let $O^k$ denote the set of pending orders during iteration $k$, with $O^0 = O$ initially, where $O$ represents all pending orders during this dispatch cycle.
And at iteration $k$,
\begin{enumerate}
  \item \textbf{Evaluation stage}: For the pending orders $O^k$ and their associated recalled courier candidates $R_o^k, {o \in O^k}$, MD scores $\{\{ f_o^r\}_{r\in R_o^k}\}_{o \in O^k}$ are calculated. 
  \item \textbf{Matching stage}: Based on current MD scores, a one(order)-to-one(courier) assignment decision is made following greedy policy (aiming to optimize the sum of MD scores for all matching relations at the current iteration). This may result in only a subset $O^k$ being successfully assigned.
  \item \textbf{Termination condition}: Denote the remaining unassigned orders as $\overline{O}^{k}$. If $\overline{O}^k = \emptyset$, stop the iterations. Otherwise, 
  update the state of couriers by including newly assigned orders, let $O^{k+1} = \overline{O}^{k}, k = k + 1$,  proceed to Step (1).
\end{enumerate}

\subsection{Order Combination Mechanism.}

Although the above algorithm has good performance in solving, it tends to promote one-to-one assignment results, which is not conducive to sufficient order pooling. To facilitate many-to-one assignments, high-quality and mutually exclusive order combinations are identified based on HPP, and incorporated into the algorithm as expanding entities rather than single orders. 
The evaluation stage at iteration $k$ is executed as follows:
\begin{enumerate}
  \item For pending orders $O^k$, calculate the HPPs between the FUs of any two orders and denote the combination set as $C^k$. Set the order combination set preserved for MD evaluation at iteration $k$ as $\widehat{C^k} = \emptyset$.
  \item Prune two-order combinations with low HPPs ($p_{o_1, o_2} < P_1$), i.e., let $C^k = C^k-C^k_{low}$.
  \item Repeat this step until $C^k = \emptyset$: pickup $c = \{o_1, o_2\} \in C^k$ with the highest HPP value, let $\widehat{C^k} = \widehat{C^k} + \{c\}$. Then remove its related entries in $C^k$ and $O^k$, i.e, let $C^k = C^k - \{c | o_1 \in c\text{ or }o_2 \in c, c \in C^k\}, O^k = O^k - \{o_1\} -\{o_2\}$.
  \item Use $\widehat{C^k}$ and $O^k$ as decision entities and calculate the MD scores with their associated couriers.
\end{enumerate}
The above process is illustrated in Figure \ref{fig:combination_details}. And in practice, $P_1$ is set to 0.6.

\subsection{Courier Recall Mechanism.}

To reduce MD score calculation volume, we can further refine the courier candidates recalled for each order/order combination using HPP.
For the evaluation stage at iteration $k$ , the pending entity sets are $\widehat{C^k}$ and $O^k$, and the courier recall mechanism is executed as follows:
\begin{enumerate}
  \item For $o \in O^k$, denote the corresponding courier candidate set as $R_o^k$. For $r \in R_o^k$, if the on-hand order set $O_r^k \neq \emptyset$, calculate the average HPP of $o$ and orders in $O_r^k$ as an estimation of MD score, i.e.  $\widetilde{f}_o^r = \frac{1}{|O_r^k |}\sum_{o' \in O_r^k} p_{o, o'}$. 
  
  For the on-hand order already picked up by courier $r$, its FU can be considered as the FU starting from the AOI where the courier is currently located and ending at its delivery AOI. 
  
  For the on-hands whose FU embedding is absent, the associated HPP is set as 0.
  
  If $\widetilde{f}_o^r$ is lower than threshold $P_2$, courier $r$ will be removed from the candidate set, i.e. $R_o^k = R_o^k - \{r\}$.
  
  \item For $c=\{o_1, o_2\} \in \widehat{C^k}$, denote the corresponding courier candidate set as $R_c^k$, which is the intersection of courier candidate sets of $o_1$ and $o_2$. For $r \in R_c^k$, calculate the average
HPP of $o_1$ and $o_2$ as Step (1), respectively. 
  
  If either $\widetilde{f}_{o_1}^r$ or $\widetilde{f}_{o_2}^r$ is lower than threshold $P_2$, courier $r$ will be removed from the candidate set, i.e. $R_c^k = R_c^k - \{r\}$.
  
  \item For orders in  $O^k$ and combinations in $\widehat{C^k}$, calculate the MD scores with their refined couriers.
\end{enumerate}
The above process is illustrated as in Figure \ref{fig:courier_recall}. And in practice, $P_2$ is set to 0.5.

\section{SEH Identification Approach} \label{APP-SEH}
We utilize BP to identify SEHs during each time interval from FUs with high FEI in a city or nearby areas. In this section, we introduce the variable definitions, objective function, and constraints of the model.

The decision variable $x_f^g$ represents whether FU $f$ belongs to SEH $g$. To calculate the average HPP in each SEH, we introduce a binary auxiliary variable $y_{f, f'} ^ g$, which indicates whether FU $f$ and $f'$ belong to SEH $g$ simultaneously.
The objective function in Equation~(\ref{eq:seh-obj}) is to maximize the average HPP in each SEH, where $p_{f, f'}$ is the HPP between FU $f$ and $f'$.

The constraint in Equation~(\ref{eq:seh-con-flow}) limits each FU to appear in only one SEH.
Equation~(\ref{eq:seh-con-flow-num}) limits the minimum and maximum number of FUs in each SEH.
Equation~(\ref{eq:seh-con-order-num}) limits the minimum number of orders in each SEH, where $n_f$ is the number of orders of FU $f$.
Equation~(\ref{eq:seh-con-y1}) and Equation~(\ref{eq:seh-con-y2}) ensure that $y_{f, f'} ^ g = 1$ if and only if $ x_{f}^g = x_{f'}^g = 1$.
Equation~(\ref{eq:seh-con-sim}) constrains the minimum average HPP in each SEH $g$.
Equation~(\ref{eq:seh-con-x}) and Equation~(\ref{eq:seh-con-y}) ensure that all the decision variables are binary.

\begin{align}
\text{max} & \sum_{g\in G} \frac{ \sum_{f\in F}\sum_{f^{'}\in F,f^{'} \neq f} p_{f,f^{'}} \times y_{f,f^{'}}^{g} } { \sum_{f\in F}\sum_{f^{'}\in F,f^{'} \neq f} y_{f,f^{'}}^{g} } \label{eq:seh-obj} \\
s.t. & \sum_{g\in\bar{G}}x_{f}^{g}=1,\forall f\in F \label{eq:seh-con-flow}  \\
&|g|^\text{min} \leq \sum_{f\in F}x_f^g \leq |g|^\text{max}\text{,}\forall g\in G \label{eq:seh-con-flow-num} \\
&\sum_{f\in F}n_f \times x_f^g \geq N,\forall g\in G \label{eq:seh-con-order-num} \\
&y_{f,f^{'}}^{g} \geq x_{f}^{g}+x_{f^{'}}^{g}-1 \label{eq:seh-con-y1} \\
&y_{f,f^{'}}^{g} \leq x_{f}^{g},y_{f,f^{'}}^{g} \leq x_{f^{'}}^{g} \label{eq:seh-con-y2} \\
&\sum_{f\in F}\sum_{f^{'}\in F,f^{'} \neq f}(p_{f,f^{'}}-P) \times y_{f,f^{'}}^{g} \geq 0,\forall g\in G \label{eq:seh-con-sim} \\
&x_f^g\in\{0,1\},\forall f\in F,\forall g\in G \label{eq:seh-con-x} \\
&y_{f,f^{'}}^{g}\in\{0,1\},\forall f,f^{'}\in F,f \neq f^{'},\forall g\in G \label{eq:seh-con-y} 
\end{align}

\balance

\end{document}